\pdfoutput=1
\documentclass[11pt]{article}

\usepackage{acl}

\usepackage{times}
\usepackage{latexsym}

\usepackage{fancyvrb}
\usepackage{graphicx}
\usepackage{multirow}

\usepackage[T1]{fontenc}
\usepackage[utf8]{inputenc}

\usepackage{microtype}

\usepackage{subcaption,booktabs}
\usepackage{makecell}
\title{\textsc{Linguist}: Language Model Instruction Tuning to Generate Annotated Utterances for Intent Classification and Slot Tagging}

\author{Andy Rosenbaum\thanks{~~Correspondence Author: <andros@amazon.com>. \newline
Author contributions are listed in Appendix \ref{sec:contributions}.} \\
  Amazon, Cambridge, USA \\
  \texttt{andros@amazon.com} \\\And
    Saleh Soltan \\
  Amazon, New York, USA \\
  \texttt{ssoltan@amazon.com}\\\And
  Wael Hamza \\
  Amazon, Dallas, USA \\
  \texttt{waelhamz@amazon.com}\\\AND
  \hspace{0.32\textwidth} Yannick Versley \\
  \hspace{0.32\textwidth} Amazon, Aachen, Germany \\
  \hspace{0.32\textwidth} \texttt{yversley@amazon.de}\\\And
  \hspace{0.32\textwidth} Markus Boese \\
  \hspace{0.32\textwidth} Amazon, Aachen, Germany \\
  \hspace{0.32\textwidth} \texttt{boesem@amazon.de} \\\And
  }

\begin{document}
\maketitle
\begin{abstract}
We present \textsc{Linguist}, a method for generating annotated data
for Intent Classification and Slot Tagging (IC+ST), via
fine-tuning \mbox{AlexaTM~5B}, 
a 5-billion-parameter multilingual
sequence-to-sequence (seq2seq) model, on a flexible instruction prompt.
In a 10-shot novel intent setting for the SNIPS dataset,
\mbox{\textsc{Linguist}}
surpasses state-of-the-art approaches
(Back-Translation and Example Extrapolation)
by a wide margin, showing absolute improvement
for the target intents of
+1.9 points on IC Recall and +2.5 points on ST F1 Score. 
In the zero-shot cross-lingual setting of the mATIS++ dataset, \textsc{Linguist} out-performs a
strong baseline of Machine Translation with Slot Alignment
by +4.14 points absolute on ST F1 Score across 6 languages,
while matching performance on IC.
Finally, we verify our results on an internal large-scale multilingual dataset for
conversational agent IC+ST and show significant improvements
over a baseline which uses Back-Translation, Paraphrasing
and Slot Catalog Resampling.
To our knowledge, we are the first to demonstrate
instruction fine-tuning of a large-scale seq2seq model
to control the outputs of multilingual intent- and slot-labeled
data generation.

\end{abstract}

\section{Introduction}
\label{sec:introduction}

\begin{figure}[h!]
\begin{Verbatim}[fontsize=\scriptsize, frame=single, commandchars=\\\{\}]
\textbf{INPUT:}
\textcolor{red}{<language>} \textcolor{teal}{\textbf{English}} \textcolor{red}{</language>}
\textcolor{red}{<intent>} \textcolor{violet}{\textbf{GetWeather}} \textcolor{red}{</intent>}
\textcolor{red}{<include>}
 \colorbox{yellow!50}{[1 \textbf{*} ]} , \colorbox{cyan!20}{[3 \underline{snow} ]} \colorbox{green!20}{[5 \emph{tomorrow} ]}
\textcolor{red}{</include>}
\textcolor{red}{<labels>}
 \colorbox{yellow!50}{[1=\textbf{geographic_poi}} , [2=country , 
 \colorbox{cyan!20}{[3=condition_description} , [4=city , 
 \colorbox{green!20}{[5=timeRange}
\textcolor{red}{</labels>}
\textcolor{red}{<examples>}
 Will the weather be okay in
  [1 Yellowstone National Park ]
  [5 one week from now ] ? \textcolor{red}{<br>}
 will it [3 rain ] at the [1 Statue of Liberty ] 
  at [5 noon ] \textcolor{red}{<br>}
 What's the weather like 
  at [1 Disneyworld ] in [5 november ] \textcolor{red}{<br>}
 I need the weather info for 
  the [1 Guggenheim Museum ] in [2 Spain ] \textcolor{red}{<br>}
  What is the weather forecast 
  for [5 October 12, 2022 ] in [4 Gyeongju ]
\textcolor{red}{</examples>}
\textbf{OUTPUTS:}

1. Give the \colorbox{yellow!50}{[1 \textbf{National Wildlife Refuge} ]} 
   forecast for \colorbox{cyan!20}{[3 \underline{snow} ]} \colorbox{green!20}{[5 \emph{tomorrow} ]}

2. Can I get the \colorbox{cyan!20}{[3 \underline{snow} ]} forecast 
   for \colorbox{yellow!50}{[1 \textbf{Lake Tahoe} ]} \colorbox{green!20}{[5 \emph{tomorrow} ]}?

3. I want to know if it will \colorbox{cyan!20}{[3 \underline{snow} ]}
   \colorbox{green!20}{[5 \emph{tomorrow} ]} at
   \colorbox{yellow!50}{[1 \textbf{Mount Rainier State Park} ]}.
\end{Verbatim}
\vspace*{-0.4cm}
	\caption{
	\textsc{Linguist} uses an instruction prompt to generate data
	with both user-requested slot values (``snow'')
	and model-generated values (``*'').
	This model \textbf{has not seen any training data for GetWeather intent, 
	or for the slot tag geographic\_poi}:
	it was fine-tuned only on the other 6 SNIPS intents.
	}
\vspace*{-0.4cm}
\label{fig:weather1}
\end{figure}

Conversational agents typically rely
on large quantities of labeled training data to understand user requests
through Intent Classification and Slot Tagging (IC+ST) \citep{TurSLU2011}. Such data is plentiful
for existing usage patterns (although costly to annotate),
yet scarce for new intents/slots and new languages. 
A growing trend to address this problem is to generate
synthetic training data, e.g. via
Paraphrasing, Back-Translation (BT), slot replacement, and Example Extrapolation (Ex2).
(\citealp{Jolly2020,Xie2020uda,zhang-etal-2020-seqmix,lee21ex2}).
In this work, we propose a novel data generation method called \textbf{L}anguage model \textbf{IN}struction tuning to \textbf{G}enerate annotated \textbf{U}tterances for \textbf{I}ntent classification and \textbf{S}lot \textbf{T}agging (\textbf{\textsc{Linguist}}).

Our \textsc{Linguist} method addresses several important gaps in the existing literature:
(1) controlling the generated data to include specific slot types and values,
(2) cross-lingual and multilingual data generation, and (3) ability to leverage the intent and slot names to inform the generation.
The key to these achievements is our design of a novel instruction prompt (Figure \ref{fig:weather1}),
consisting of natural language descriptions for the desired model outputs.
We first fine-tune a large pre-trained seq2seq Transformer \citep{NIPS2017_3f5ee243} model to learn 
how to generate annotated utterances following the prompt instructions.
Then, for a novel intent or slot with only a few or even zero training examples,
we apply the model to generate similar data, which we add to the training set for
an IC+ST model.

We demonstrate the effectiveness of \textsc{Linguist} on three datasets by showing
substantial improvements over strong baselines.
(i) On a 10-shot novel intent setting with English SNIPS \citep{Coucke18snips}, \textsc{Linguist} improves over Back-Translation and Ex2 by +1.9 points absolute on IC and +2.5 points absolute on ST.
(ii) On cross-lingual mATIS++ \citep{xu20matis}, \textsc{Linguist} out-performs
the best Machine Translation plus slot alignment reported by \citeauthor{xu20matis}, 
by +4.14 points in ST across 6 languages, while matching performance on IC.
(iii) Finally, to demonstrate the success of our method on a real-world conversational agent system,
we apply \textsc{Linguist} on an internal dataset containing hundreds of intents and slot types across 4 languages,
and show large improvements over a baseline which uses Back-Translation and Paraphrasing.

We also show \textsc{Linguist} can generate
IC+ST-annotated data from \textit{zero examples},
using only the natural language intent and slot label names.
\mbox{\textsc{Linguist}} achieves 80.0 IC Recall and 56.9 ST~F1 Score on
new SNIPS intents despite never seeing a single example for the new intents.
To our knowledge, \textsc{Linguist}
is the first system capable of generating
IC+ST-annotated data in this setting.

\section{Related Work} % (fold)
\label{sec:related_work}

Large-scale Language Models (LLMs) such as GPT \citep{Radford19gpt2,Brown2020gpt3}
and AlexaTM 20B \citep{Soltan2022AlexaTM2F} excel at 
performing novel tasks with only a few examples
via in-context learning,
i.e. without requiring any model parameter updates.
For example, \citet{sahu22ots} apply GPT-3 to generate variations of examples from a given class.
\citet{Wang21UDG} generate both text and class label together via GPT-3,
towards eliminating the need for human labeling.

Pre-training then fine-tuning of seq2seq models was introduced in English BART and multilingual mBART \citep{lewis20bart,liu20mbart}.
T5 and mT5 \citep{raffel20t5,xue-etal-2021-mt5}
extended the idea by framing more downstream tasks as text-to-text.
FLAN \citep{wei2022finetuned} introduced \textbf{instruction tuning},
where a large-scale seq2seq model is fine-tuned on instruction prompts
from a variety of tasks,
in order to generalize to new tasks without any further parameter updates.

Prior work also explores conditioning generation on intent and slot labels.
\citet{ding-etal-2020-daga} train a conditional language model on
a mixture of annotated and unannotated text, allowing to sample novel annotated utterances.
\citet{malandrakis-etal-2019-controlled} train a seq2seq model
from interpretation-text pairs, applying
variational auto-encoders for more diversity.
\citet{Jolly2020} expand on this, exploring different sampling
strategies, adding more variety by shuffling slot names, and examining the behavior where a new intent is introduced with limited training data. \citet{convai21-paraphrase-generation} extend this to the multilingual setting.
Generative Insertion Transformers \citep{kumar-etal-2022-controlled}
generate carrier phrases for a target intent and containing specific entities.
A limitation of all these approaches is that the trained model cannot generalize to 
novel intents and slots at inference time.

Generative Conversational Networks \citep{papangelis21gcn} are trained via reinforcement learning to
generate annotated data from seed examples, studying English IC+ST and other tasks.

The closest relative of \textsc{Linguist} is Example Extrapoloation (Ex2) \citep{lee21ex2},
which generates annotated IC+ST data using a seq2seq model and provided seed examples.
We compare \textsc{Linguist} and \textsc{Ex2} in more detail in section \ref{sec:comparison_ex2}.

A widely used paraphrasing method
is \textbf{Back-Translation} (BT), i.e.
translating text from one language into another ``pivot'' language,
then back again. \citet{Bannard2005}
extract paraphrases directly from parallel corpora.
\citet{sennrich-etal-2016-improving} and 
\citet{edunov-etal-2018-understanding} use BT for Machine Translation
and \citet{Xie2020uda} for data augmentation on classification tasks.

Other approaches directly target the Paraphrasing task: \citet{Prakash2016}
learns an LSTM model by supervised training on a paraphrase corpus,
whereas \citet{kumar-etal-2020-data} use an unsupervised denoising task, in
both cases only using text and not covering slot labels.
\citet{cho-etal-2019-paraphrase} explore Paraphrasing via Semi-Supervised Learning.

A different approach to data augmentation is token replacement:
SeqMix \citep{zhang-etal-2020-seqmix} replaces tokens with the nearest
neighbor in the embedding space, \citet{dai-adel-2020-analysis} replaces slots
with synonyms, or mentions from other instances of the same label.
Easy Data Augmentation \citep{wei-zou-2019-eda}
includes synonym replacement, random insertion, random swap, and random deletion for 
classification task.
\citet{zheng2021fewnlu} benchmark the success of LLMs on few-shot settings.

\section{\textsc{Linguist} Data Generator Model}
\label{sec:linguist-model}

\textsc{Linguist} provides three key innovations compared to prior data generation work:
(1) controls for slot types and values (either user-supplied or model-generated) to include in the outputs;
(2) multilingual and cross-lingual generation; and
(3) ability to leverage the natural language label names to inform generation, enabling a new ``Label Names Only''
setting (Section \ref{sec:nifs_lno}).

\subsection{\textsc{Linguist} Prompt Design} % (fold)
\label{sub:linguist_prompt_design}

We control the generation output via a novel
prompting scheme, as shown in Figure \ref{fig:weather1}.
The prompt contains five blocks:
(i) the output \texttt{<language>},
(ii) the \texttt{<intent>} name,
(iii) which slot types and values to \texttt{<include>} in the output,
(iv) a mapping from \texttt{<labels>} to numbers, and
(v) up to 10 \texttt{<examples>}, each belonging to the same intent, and including
zero or more of the available labels.

The \texttt{<include>} block instructs \textsc{Linguist} which slot types and values to generate,
such as \texttt{[3 snow ]}, where the number corresponds to the \textbf{slot type} (in this case, \texttt{[3=condition\_description}),
and the content inside the brackets is the value to use for that slot (here ``snow'').
To increase the diversity of outputs, the user may also instruct the model to generate a value for a slot, 
using the \textbf{``wildcard'' token}, e.g. \texttt{[1 * ]} indicates that the model should
sample a value for slot number 1.

As shown in Figure \ref{fig:weather1}, \textsc{Linguist} learns to produce a rich sample of values 
even for intents and slot types it never saw during fine-tuning. For example, in this case
the wildcard is for \texttt{[1=geographic\_poi} and the model outputs sensible values such as
``Lake Tahoe'', despite this phrase never appearing the fine-tuning data.

\subsection{Training the \textsc{Linguist} model} % (fold)
\label{sub:training_the_linguist_model}

We fine-tune a pre-trained seq2seq model on pairs of \textsc{Linguist} instruction prompts
and corresponding annotated target utterances, derived from a
de-duplicated IC+ST task dataset $R$.
Specifically, we format an instruction prompt $p_i$ targeting each utterance $t_i \in R$,
including in $p_i$ up to 10 \emph{other} example utterances
$E = \{e_j\}_{j=1}^{10} \in R ~\textrm{s.t.}~ \forall j,~e_j \neq t_i~\textrm{and}~\textrm{intent}(e_j) = \textrm{intent}(t_i)$.
To make the generation robust to the number of provided examples, we do not always include all 10 in the prompt,
but instead randomly select $k$ examples from $E$, with $k$ chosen randomly between 0 and 10, or the number of utterances available that share the same intent as $t_i$, whichever is smaller.
We never duplicate utterances in the prompt. Finally, we produce a corpus of training prompts equal in size to the original IC+ST training set.

To reduce the tendency for the model to overfit on the intent and slot labels
(as observed by \citealp{lee21ex2}),
we drop out the label names for both at a rate of 0.2,
replacing the label name e.g. \texttt{GetWeather} with a random sequence of
between 1 and 5 letters like \texttt{A\_Q\_Y}.
(Ablation in Appendix \ref{sec:ablation_label_names_dropout}.)
The intuition for masking the labels rather than skipping the
\texttt{<intent>} and \texttt{<labels>} blocks
is to provide the model a consistent signal for position embeddings,
and always allowing it to attend to these tags
if it wishes to.

\label{ssub:learning_to_generate_slot_values}

% (Learning the star)
To jointly teach the model both to copy user-supplied slot values like
\texttt{[3 snow ]} and to produce appropriate values for the wildcard \texttt{[1 * ]},
we format the training prompts with examples of both.
For the prompt $p_i$ targeting utterance $t_i$, we randomly select from a Geometric distribution
$d \sim \textrm{Geo}(0.5)$ ($0 \leq d \leq \textrm{\# slots in}~t_i$) slots and 
replace their values with \texttt{"*"} in the \texttt{<include>} block of the prompt.
The effect is that approximately 50\% of utterances have all slot values replaced by the wildcard token,
25\% of utterances keep one slot value, etc.

We do not add the tags \texttt{<intent>}, \texttt{[1}, etc. to the model's sentencepice \citep{kudo-richardson-2018-sentencepiece}
tokenizer vocabulary (Appendix \ref{sec:tokenizer_choices}).

\subsection{Comparing \textsc{Linguist} to \textsc{Ex2}}
\label{sec:comparison_ex2}
The closest relative to our approach is Example Extrapolation (Ex2, \citealp{lee21ex2}),
which also produces
slot-labeled text from seed examples. The
novelty of \textsc{Linguist} compared to Ex2 is threefold: (i) instructions to control the slot types and values generated,
(ii) multilingual and cross-lingual, and (iii) the ability to include label and slot names
in the prompt.
In particular, 
\textsc{Ex2} showed that \emph{anonymizing} the labels
improved on IC however hurt ST.
Our \textsc{Linguist} model improves over our implementation of Ex2 on both IC and ST,
and furthermore the labels enable \textsc{Linguist} to
perform one-shot and zero-shot for new intents and slots,
as we show in Section \ref{ssub:linguist_for_snips_lno} and Appendix~\ref{sec:sample_model_outputs}.

\section{Experimental Setup}
\label{sec:experiments}

This section describes the datasets, tasks, IC+ST model, baseline gata generation methods, and metrics that we
use to evaluate \textsc{Linguist}.

\subsection{Datasets} % (fold)
\label{sub:datasets}

\subsubsection{SNIPS Dataset} % (fold)
\label{ssub:snips_dataset}

The SNIPS dataset
\citep{Coucke18snips} is a public IC+ST benchmark consisting of 7 intents, each with between 2 and 14 slot types (39 unique slot types in total). It includes around 2k training utterances and 100 validation utterances per intent.
In order to avoid overfitting our method on the small validation set, at the beginning of our experiments,
we partition the training set into 97\% Train and 3\% Development.
We use our Development set split for iterating on all modeling and data processing decisions,
including the hyperparameters for \textsc{Linguist} and hyperparameters and selection of best checkpoint
for the encoder fine-tuning on IC+ST.
Only at the very end of our experiments, we evaluate and report on the Validation set.
See Table \ref{tab:snips-data-counts} for counts of Train/Dev/Valid utterances.

\begin{table}[h]

\centering
\small
\begin{tabular}{lccc}
\hline
 Intent               &   Train &   Dev. &   Valid. \\
\hline
 AddToPlaylist        &    1884 &            58 &          100 \\
 BookRestaurant       &    1914 &            59 &          100 \\
 GetWeather           &    1940 &            60 &          100 \\
 PlayMusic            &    1940 &            60 &          100 \\
 RateBook             &    1898 &            58 &          100 \\
 SearchCreativeWork   &    1896 &            58 &          100 \\
 SearchScreeningEvent &    1901 &            58 &          100 \\
\hline
 Total                &   13373 &           411 &          700 \\
\hline
\end{tabular}

\caption{Data counts per intent for SNIPS.}

\label{tab:snips-data-counts}
\end{table}

\subsubsection{Multilingual ATIS++} % (fold)
\label{ssub:multilingual_atis}

For cross-lingual experiments we evaluate on mATIS++
\citep{xu20matis}, which consists of human-translated text and annotations from the original English travel information requests
ATIS dataset \citep{Hemphill1990TheAS} plus Hindi and Turkish translations from mATIS \citep{Upadhyay2018AlmostZC}.
Our experiments cover the 7 languages that mATIS++ shares with our
pretrained model:
English, Spanish, German, French, Portuguese, Japanese, and Hindi, with 4488 (HI: 1440)
training utterances covering 18 (HI: 17) intents and 84 (HI: 75) slots.

To demonstrate the cross-domain adaptation of the \textsc{Linguist} method, we use
the \textbf{MASSIVE} dataset \citep{jgmf22massive} covering 51 languages with parallel versions
of the (English-only) SLU Resource Package \citep{bastianelli-etal-2020-slurp} utterances,
covering 20k utterances per language across 18 domains, 60
intents and 55 slots.
We use MASSIVE only to train a \textsc{Linguist} model, then apply the model to mATIS++.
In order to keep mATIS++ as a novel domain, we exclude the somewhat related \texttt{transport} domain from MASSIVE when
we train the \textsc{Linguist} model.

\subsubsection{Internal Dataset} % (fold)

\label{ssub:internal_dataset}

To demonstrate the value of our method to a real-world setting, we benchmark on an internal large-scale
multilingual dataset representative of requests to
a conversational agent.
We consider five portions of the dataset, known as \emph{features}, namely:
CameraControl, ClockSettings, HomeSecurity, Music, and Timers, each containing
one or more intents, and one or more associated slots.
For each feature, there is a ``starter'' training set comprised of a few dozens of annotated utterances which were curated for the new feature, and a test set containing hundreds of annotated utterances pertaining to that new feature.
Additionally, there is a large training dataset $E$ of annotated utterances from existing features.
\emph{The Existing Features training data $E$  does not contain examples of any of the new features.}

\subsection{Evaluation Tasks}

\subsubsection{New-Intent Few-Shot (NIFS)} % (fold)
\label{ssub:new_intent_few_shot_nifs}

\begin{figure}[h!]
\centerline{\includegraphics[width=0.48\textwidth]{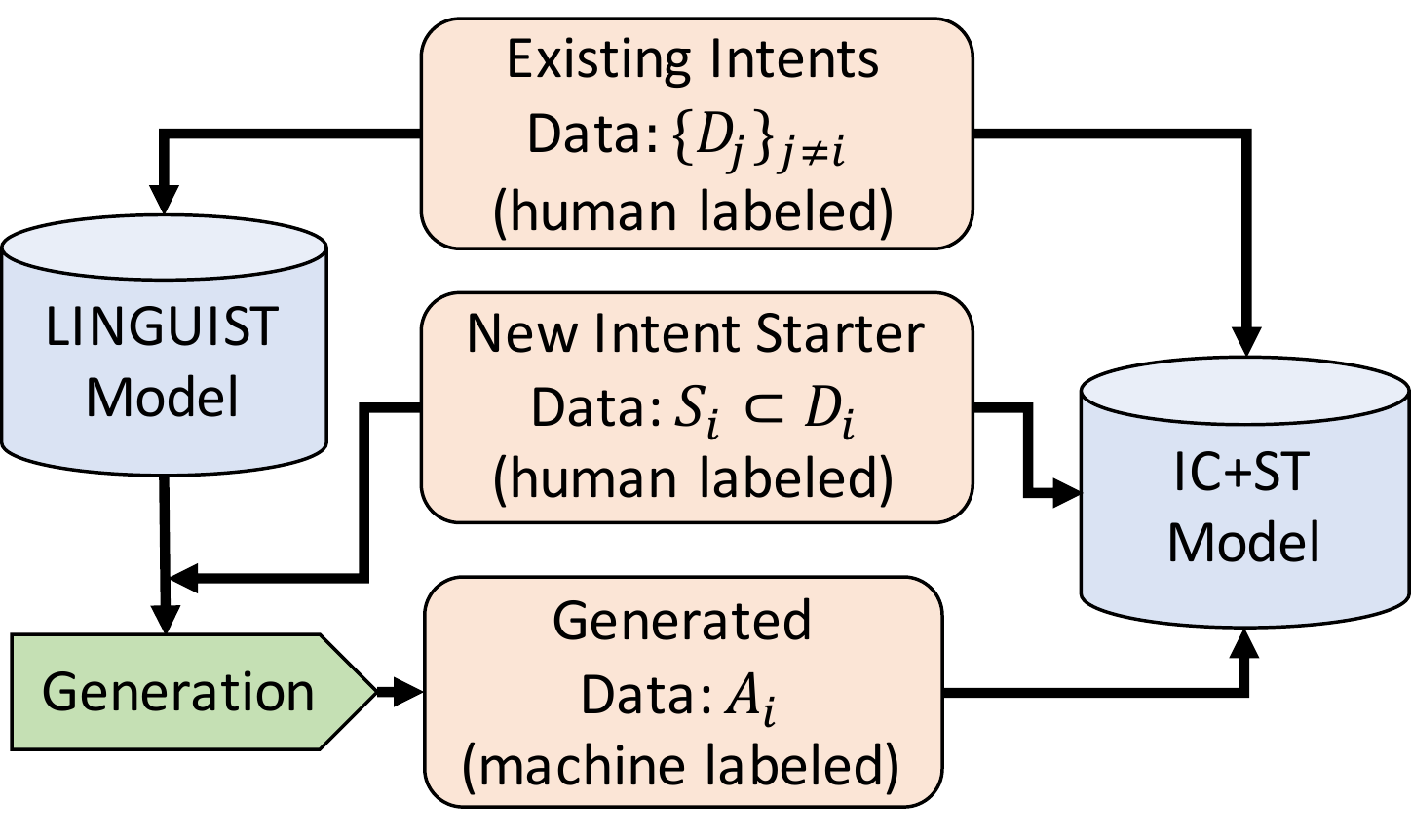}}

\caption{New-Intent Few-Shot (NIFS) Setup.}
\label{fig:setup}
\end{figure}

As shown in Figure \ref{fig:setup}, we simulate
the introduction of a new intent
into an existing well-resourced dataset.
Given a training dataset $R~=~\bigcup{}_{j=1}^{m} D_j$ containing data $D_j$ for $m$
intents $j=1 \dots m$,
we select an intent $i \in \{ 1, \dots, m \} $, and reduce its training data to
only a small number $K$ of ``starter'' utterances $S_i \subset D_i$.
We apply various data augmentation techniques on $S_i$
to create augmented data $A_i$.
Finally, we train an IC+ST model using 
$R_i' = S_i \cup A_i \cup \{ D_j \}_{j \neq i} $,
i.e. the concatenation of starter and augmented data for intent $i$
with the unmodified data for all other intents.

The internal dataset
is already split in this way, however at the \emph{feature} rather than intent level.

% Description of training process
For SNIPS, we create 7 NIFS settings, one for each intent,
reducing its training data down to only $K$=10 starter utterances $S_i$.
We create 5 versions, each with a different random seed
for selecting $S_i$,
and always including at least one example for all slot types that occur for intent $i$.

To demonstrate the ability of \textsc{Linguist} to generalize to new intents and slots at inference time,
\textbf{we exclude the new intent's starter utterances from fine-tuning}.
For each intent $i$, we train a \textsc{Linguist}
model on the other 6 intents $\{ D_j \}_{j \neq i}$.
Then, \emph{during inference},
we forumlate prompts with the starter utterances between
\texttt{<example>} and \texttt{</example>},
and generate more data.
Note, this generation step \textbf{does not require any model parameter updates}.

\subsubsection{NIFS Label Names Only (LNO)}
\label{sec:nifs_lno}

In this more challenging variant of NIFS,
\emph{only the intent and slot label names} are available for the starter utterances,
not their text or annotation.
This is useful when developing new
intents as we need only specify which slot types
can go together, and need not curate or annotate 
any real examples.
Notably, to the best of our knowledge,
\textbf{\textsc{Linguist} is the first system capable of generating intent- and slot-annotated data in this setting}
(as shown in Figure \ref{fig:output_lno}, Appendix \ref{sec:example_lno}), 
by attending to the natural language label names in the prompt.

\subsubsection{Zero-Shot Cross-Lingual} % (fold)
\label{ssub:cross_lingual_zero_shot}

For mATIS++, we evaluate in the \emph{zero-shot cross-lingual} setting, where real training data
is available only for English.
We fine-tune an IC+ST model on the English training data plus any augmented
examples generated from this data, and evaluate the model on the test sets from all languages.

\subsection{Models}

Our experiments rely on two pre-trained models:
(1) \textbf{AlexaTM 5B} (described next in Section \ref{sub:data_gen_mdoel}) which we fine-tune on the \textsc{Linguist}
prompts and use it to generate IC+ST training data, and
(2) \textbf{xlm-roberta-base} which we fine-tune on the IC+ST task
including the data generated from \textsc{Linguist}.

\subsubsection{AlexaTM 5B}
\label{sub:data_gen_mdoel}

AlexaTM 5B is a multilingual seq2seq Transformer \citep{NIPS2017_3f5ee243}
model pre-trained similar to AlexaTM 20B \citep{Soltan2022AlexaTM2F},
however with denoising objective only (i.e. without Causal Language Modeling objective).
Like \mbox{AlexaTM~20B}, the architecture is derived from the HuggingFace \citep{wolf-etal-2020-transformers} BART 
\citep{lewis-etal-2020-bart} class, and consists of 29 encoder and 24 decoder layers,
hidden dimension 2560, and 32 attention heads.
The model is trained on 900B tokens of Wikipedia and mC4 \citep{xue-etal-2021-mt5} of 12 languages as used in AlexaTM 20B. We used a maximum sequence length of 512 and a batch size of 1M tokens.
The encoder weights of AlexaTM 5B are initialized with the 2.3B-parameter Alexa Teacher Model encoder \citep{FitzGerald2022} trained on MLM task on the same data, frozen during the first half of the AlexaTM 5B pre-training.
Hyper-parameters used for fine-tuning AlexaTM 5B on \textsc{Linguist} are described in Appendix \ref{sec:linguits_training_details}.

\subsubsection{IC+ST Fine-tuning} % (fold)
\label{sub:ic_st_model_training}

For SNIPS and mATIS++, following \citet{chen2019bert}, we fine-tune a BERT-style model for joint IC+ST.
On top of the encoder hidden states, we attach two separate classification heads, one for IC and another for ST.
Each head consists of two layers of 256 hidden dimension, with gelu activation, dropout 0.2, and layer norm.
The IC head utilizes representation from the first token of the sequence (\texttt{[CLS]}), while the ST head utilizes the 
first subword token of each word.

For our encoder, we use \texttt{xlm-roberta-base} \citep{conneau-etal-2020-unsupervised} (12 layers, 768 hidden dimension),
from the HuggingFace \citep{wolf-etal-2020-transformers} implementation.
We fine-tune with batch size 128 for 3k updates (i.e. 30 epochs for the full-size data).
We freeze the embedding layer; all other parameters are free to update during training.
We use Adam \citep{Kingma2015AdamAM} with peak learning rate 3e-5, increased linearly from 0 to 600 updates, then decayed linearly to 0 until the end of training.

To avoid over-fitting on the official SNIPS Validation dataset, we use our Development split (Section \ref{ssub:snips_dataset}) for early stopping, 
selecting the checkpoint with best performance on ST. All of our IC+ST fine-tuning runs for SNIPS use identical hyper-parameters, regardless of the data
generation method being explored.
For each data generaiton method, we train and test 7 different Joint IC+ST models $\{ {M_i} \}_{i=1}^{7}$ in NIFS setting:
each using a combination of the modified data for intent $i$, and the unmodified data for all other intents.

For mATIS++, we follow the same model architecture settings and train for 2k updates
(64 epochs for English only data, or 9 epochs when using data from all the 7 languages.)
We select the checkpoint with best ST F1 Score on the English dev set only.

For our internal benchmark, we use similar settings,
however with a smaller internal Transformer-based encoder for fine-tuning on the IC+ST task.

\subsection{Baseline Data Generation Methods}
\label{sub:baselines}

% subsubsection baseline_methods_for_snips (end)

The \textbf{Interpretation-Conditioned Language Model} (ICLM) \citet{Jolly2020}
generates unlabeled text conditioned on intent and provided slot values, 
with a separate label projection step to recover the full slot annotation.
ICLM does not generate novel slot values.
Our implementation uses a small Transformer \citep{NIPS2017_3f5ee243} architecture with $\sim$~37M parameters,
and a simple character-level Levenshtein distance measure to project the slot labels.
We produce 50 outputs per input, then filter/de-duplicate (see Appendix \ref{sec:filtering_iclm_outputs}).

We apply \textbf{Back-Translation} (BT)
using two separate MT systems to show the influence of the translation model.
The first uses the open-source Sockeye toolkit \citep{hieber-etal-2018-sockeye}
and a small (91M parameters) Transformer
which has been fine-tuned on around 10k utterances of annotated parallel data.
We use fast\_align \citep{dyer-etal-2013-simple} to project the slot labels
to the generated utterances.
We call this system \textbf{``BT-Small''}. We use M=1 forward and N=10 backward
translations to obtain 10 paraphrases, and then filter and deduplicate
(see Appendix \ref{sec:filtering_bt_small}). We use French (SNIPS) or English (Internal) respectively
as pivot languages.

For a stronger BT baseline ``\textbf{BT-5B}'', we build an MT system
by fine-tuning AlexaTM 5B on  WMT14
(retrieved from HuggingFace datasets)
jointly on en$\rightarrow$fr and fr$\rightarrow$en using an instruction prompt
(prefixing the input text with \texttt{Translate to French:} or \texttt{Translate to English:}, respectively)
to control the translation direction.
We use SimAlign \citep{jalili-sabet-etal-2020-simalign} to project the slot labels to the paraphrased text.
For SNIPS, we use French as the pivot language, with beam search 10 both forward and backward,
producing 100 outputs per original sentence, then filter and de-duplicate the outputs (Appendix \ref{sec:filtering_bt_5b_outputs}).
BT-5B was not available for the Internal Benchmark.

For SNIPS, we implement \textbf{Example Extrapolation} (Ex2, \citet{lee21ex2}) with the default ``fully anonymized labels''
setting, again fine-tuning from AlexaTM 5B,
training a separate version for each intent's experiment, as described in
\ref{ssub:new_intent_few_shot_nifs}.

\textbf{Slot Catalog Resampling} is a simple approach to data augmentation which samples entities from a catalog for a particular label.
For example, given an utterance like ``play jason mraz''
we might sample ``weeezer'' from a catalog of artist names, to get 
``play weezer''.
We only use Slot Catalog Resampling for the Internal Benchmark, as there are no slot catalogs 
available for SNIPS or mATIS++.

\subsection{Metrics} % (fold)
\label{sub:metrics}

\subsubsection{Metrics for SNIPS and mATIS++} % (fold)
\label{ssub:metrics_for_snips}

We use separate metrics to measure (1) support for the 
new intent, while (2) not harming the overall performance across all intents.
For (1), we run the model on a test set containing \emph{only} the new intent.
We refer to this as the \emph{Local} Intent Recall (IR), and \emph{Local} ST F1 Score.
To measure (2), we run the model on the combined test set of all intents together,
and call this the \emph{Global} Intent Accuracy (IA) and \emph{Global} ST F1 Score.
In both cases, for ST F1 Score, we ignore the ``O'' (non-entity) tag,
using the seqeval \citep{seqeval} implementation.

When training data is modified for a particular intent $i$, the Local
metrics for $i$ change across methods as expected, whereas changes in Global metrics
(see Appendix \ref{sec:results_on_global_metrics}) are very small for all methods.

For the cross-lingual \textbf{mATIS++} experiments, we report (Global) intent accuracy and
Slot F1, since we are doing cross-lingual transfer for the whole dataset, and not targeting specific intents.

\subsubsection{Metrics for Internal Benchmark} % (fold)
\label{ssub:metrics_for_internal_benchmark}

For the internal benchmark, we only evaluate in the \emph{Local} setting. 
We measure Semantic Error Rate (SemER: \citealp{Su2018ARS} or Appendix \ref{sec:semer_metric}) which
jointly evaluates the IC and ST performance.
Lower SemER indicates improvement to the system.
We report relative reduction in SemER, where a negative number indicates improvement.

\begin{table*}[h]
\setlength{\tabcolsep}{3.7pt}
\begin{subtable}[h]{1\textwidth}
\centering
\small
\makebox[\textwidth][c]{
\begin{tabular}{lcc | c | cccc | c}
\hline
 Modified Intent / Data   &   Full & s10-NoUps        & \thead{s10}      & \thead{s10 \\ +ICLM}   & \thead{s10 \\ +BT-Small}   & \thead{s10 \\ +BT-5B}    & \thead {s10 \\ +Ex2}   & \thead{s10 \\ +\textbf{\textsc{Linguist}}}   \\
\hline
 AddToPlaylist            &  100.0 & 95.6 $\pm$6.8  & \underline{98.3} $\pm$2.1 & 97.5 $\pm$2.2   & \underline{98.3} $\pm$2.4       & \textbf{99.4} $\pm$0.5  & 97.5 $\pm$1.6  & 93.9 $\pm$3.2       \\
 BookRestaurant           &  100.0 & 91.0 $\pm$3.4  & 93.5 $\pm$2.4 & \underline{93.8} $\pm$1.8   & 92.5 $\pm$1.7       & \textbf{94.6} $\pm$1.1  & 91.3 $\pm$5.4  & \textbf{94.6} $\pm$1.5       \\
 GetWeather               &  100.0 & 98.8 $\pm$0.4  & 98.8 $\pm$0.4 & 99.4 $\pm$0.5   & 99.6 $\pm$0.5       & \underline{99.8} $\pm$0.4  & \underline{99.8} $\pm$0.4  & \textbf{100.0} $\pm$0.0      \\
 PlayMusic                &   99.0 & 70.8 $\pm$10.1 & 77.1 $\pm$6.6 & 79.8 $\pm$8.4   & 76.5 $\pm$5.8       & \underline{84.0} $\pm$2.8  & 83.8 $\pm$7.5  & \textbf{90.4} $\pm$4.7       \\
 RateBook                 &  100.0 & 99.0 $\pm$0.0  & 99.6 $\pm$0.5 & \textbf{100.0} $\pm$0.0  & \underline{99.8} $\pm$0.4       & 99.6 $\pm$0.5  & \underline{99.8} $\pm$0.4  & \textbf{100.0} $\pm$0.0      \\
 SearchCreativeWork       &  100.0 & 69.0 $\pm$8.8  & 76.9 $\pm$9.3 & 74.2 $\pm$9.1   & 73.3 $\pm$13.6      & 79.4 $\pm$11.3 & \underline{80.8} $\pm$3.9  & \textbf{83.3} $\pm$6.9       \\
 SearchScreeningEvent     &   95.2 & 66.5 $\pm$5.6  & 73.1 $\pm$8.2 & 72.1 $\pm$3.7   & 71.5 $\pm$6.1       & 73.5 $\pm$10.8 & \underline{77.3} $\pm$9.1  & \textbf{81.9} $\pm$3.9       \\
\hline
 Average                  &   99.2 & 84.4 $\pm$2.9  & 88.2 $\pm$1.9 & 88.1 $\pm$2.4   & 87.4 $\pm$2.9       & \underline{90.1} $\pm$1.6  & 90.0 $\pm$2.4  & \textbf{92.0} $\pm$0.8       \\
\hline
\end{tabular}
}

\caption{SNIPS New-Intent Few-Shot (NIFS) results on \textbf{Local Intent Recall}.}

\label{tab:snips-results-new-ic}
\end{subtable}

\hfill

\begin{subtable}[h]{1\textwidth}
\centering
\small
\makebox[\textwidth][c]{
\begin{tabular}{lcc | c | cccc | c}
\hline
 Modified Intent / Data   &   Full & s10-NoUps        & \thead{s10}      & \thead{s10 \\ +ICLM}   & \thead{s10 \\ +BT-Small}   & \thead{s10 \\ +BT-5B}    & \thead {s10 \\ +Ex2}   & \thead{s10 \\ +\textbf{\textsc{Linguist}}}   \\
\hline
 AddToPlaylist            &   94.1 & 76.8 $\pm$2.9  & 81.2 $\pm$2.5  & 78.4 $\pm$2.4   & \textbf{82.0} $\pm$1.8       & \underline{81.3} $\pm$1.7  & 80.6 $\pm$3.1  & 80.9 $\pm$3.4       \\
 BookRestaurant           &   96.4 & 71.9 $\pm$2.3  & 81.3 $\pm$2.1  & 80.4 $\pm$1.4   & 81.2 $\pm$1.2       & \underline{83.3} $\pm$2.5  & 78.6 $\pm$4.5  & \textbf{83.4} $\pm$1.7       \\
 GetWeather               &   97.8 & 74.7 $\pm$3.9  & \underline{84.9} $\pm$5.4  & 82.9 $\pm$4.8   & 82.3 $\pm$4.5       & 84.0 $\pm$2.8  & 82.9 $\pm$4.2  & \textbf{85.4} $\pm$2.8       \\
 PlayMusic                &   91.7 & 42.0 $\pm$4.3  & 59.2 $\pm$2.2  & 58.0 $\pm$3.4   & 56.1 $\pm$3.1       & 65.4 $\pm$4.2  & \underline{67.6} $\pm$6.2  & \textbf{70.1} $\pm$1.8       \\
 RateBook                 &   99.7 & 89.4 $\pm$1.5  & \underline{95.0} $\pm$0.9  & \textbf{95.4} $\pm$0.8   & 93.5 $\pm$3.3       & 93.6 $\pm$1.5  & 94.7 $\pm$0.6  & 94.8 $\pm$1.7       \\
 SearchCreativeWork       &  100.0 & 56.2 $\pm$10.5 & 70.9 $\pm$11.3 & 67.6 $\pm$9.8   & 68.9 $\pm$11.1      & 72.9 $\pm$12.1 & \underline{75.2} $\pm$5.2  & \textbf{79.3} $\pm$5.0       \\
 SearchScreeningEvent     &   96.6 & 56.4 $\pm$7.4  & 71.6 $\pm$3.6  & 72.8 $\pm$4.6   & 69.8 $\pm$4.6       & 74.2 $\pm$3.6  & \underline{79.0} $\pm$4.3  & \textbf{82.3} $\pm$3.4       \\
\hline
 Average                  &   96.6 & 66.8 $\pm$2.2  & 77.7 $\pm$2.0  & 76.5 $\pm$2.1   & 76.3 $\pm$2.5       & 79.2 $\pm$2.8  & \underline{79.8} $\pm$2.1  & \textbf{82.3} $\pm$1.3       \\
\hline
\end{tabular}
}
\caption{SNIPS New-Intent Few-Shot (NIFS) results on \textbf{Local ST F1 Score}.}
\label{tab:snips-results-new-ner}
\end{subtable}

\hfill

\label{tab:local_results}
\caption{Our main results on SNIPS Validation set (Section \ref{ssub:snips_dataset}).
For each cell $(i, j)$, we train a joint IC+ST encoder on the combination of
data from intent $i$ modified according to strategy $j$, and all other intents' data unmodified.
``Full'' is trained on the full dataset without any modifications;
for ``s10-NoUps'', the data for intent $i$ is reduced to only 10 ``starter'' examples, and
are \emph{Not Up-sampled};
for ``s10'', the starter utterances are up-sampled to $N_i$, the original
data size for intent $i$.
For the remaining columns, the up-sampled starter utterances for intent $i$ are mixed with augmented data
derived from them using a particular method,
which is re-sampled to $N_i$ in size.
``s10+X'' uses ICLM, BT-Small, BT-5B, respectively.
``s10+Ex2'' uses our internal 5B seq2seq model with Ex2,
``s10+\textsc{Linguist}'' uses data generated by our \textsc{Linguist} method.
We bold (underline) the mean for the method with best (second best) results.
Experiments are run across five random seeds, as mean $\pm$ standard deviation.}
\end{table*}

\section{Results} % (fold)
\label{sec:results}

% subsection evaluation (end)
\subsection{SNIPS Results}
\label{sub:snips_exp}

The main results are presented in
Table \ref{tab:snips-results-new-ic} for Local Intent Recall and
Table \ref{tab:snips-results-new-ner} for Local ST F1 Score.

% subsection subsection_name (end)
\subsubsection{Baseline Results on SNIPS} % (fold)
\label{ssub:baseline_results_for_snips}

An upper bound for the New-Intent Few-Shot (NIFS) setting,
is a model trained on the full dataset,
which we train and report (``Full'' in the tables) at 99.2 for Local Intent Recall and 96.6 for Local ST F1 Score.
Reducing to 10 utterances (``s10-NoUps'') harms both IC and ST,
(although ST more substantially),
however simply up-sampling (duplicating)
the starter utterances (``s10'') recovers a sizeable portion of the
performance lost.

The rest of the columns use a mix (weighted 0.5/0.5) of the up-sampled 10 starter utterances,
plus augmented data derived from them via the specified methods.
In all cases, we re-sample the final amount of data for the target intent to match the count in the original unmodified dataset.

We find that ICLM and BT-Small do not improve on
Local Intent Recall or Local ST F1 Score compared to ``s10'', whereas
BT-5B is a strong baseline, achieving 90.1 vs 88.2 for IC and 79.2 vs 77.7 for ST.
Compared to BT-5B, Ex2 matches for IC at 90.0 and only slightly improves for ST at 79.8.

\subsubsection{\textsc{Linguist} Results on SNIPS} % (fold)
\label{ssub:linguist_for_snips}

We train 7 versions of the \textsc{Linguist} model
one for each heldout intent, as described in section \ref{ssub:new_intent_few_shot_nifs}.

Utilizing the ability of \textsc{Linguist} to both copy slot values and produce novel values,
we format multiple prompt versions ${p_i}_k$ from each starter utterance $s_i$.
The first, dubbed ``copy-all'' instructs \textsc{Linguist} to copy all the slot values,
while producing new carrier phrases. Note that \mbox{\textsc{Linguist}} may also re-order the slots in the sentence.

Then, for each slot type $k$,
we create a new version of the prompt replacing the value for $k$
with the wildcard \texttt{"*"},
instructing \textsc{Linguist} to produce a new value for the slot,
while copying the other slot values as they are, and generating a suitable carrier phrase.
We refer to this strategy as ``sample-each''.
We use \emph{top\_$k$} sampling with $k=50$ and temperature 0.3 to generate 
100 utterances per prompt.

% Mixing
After filtering the generated utterances (see Appendix~\ref{sec:filtering_linguist_outputs} for details),
we mix the up-sampled 10 starter utterances with the \textsc{Linguist}-generated data.
We use identical settings for \textsc{Linguist} fine-tuning and generation across all 
35 runs (7 intents times 5 random seeds) for the SNIPS-NIFS benchmark.
Following the setting of the other baselines, we fine-tune the IC+ST model
on the concatenation of the augmented and mixed data for intent $i$
with the original data for all other intents.

Compared to Ex2 (``s10+Ex2''),
\textsc{Linguist} improves by
\textbf{+2.0 points absolute on Local Intent Recall} (from 90.0 to 92.0), and
\textbf{+2.5 points absolute on Local ST F1 Score} (from 79.8 to 82.3).

Finally, we show that the improvements in Local metrics for the new intent
do not cause harm to the overall system, and in fact provide a small improvement.
As shown in Table \ref{tab:snips-results-overall-ic} and 
Table \ref{tab:snips-results-overall-ner} (Appendix \ref{sec:results_on_global_metrics_nifs}),
``s10+\textsc{Linguist}'' improves upon ``s10+Ex2''
by +0.3 points absolute on both Global Intent Accuracy and Global Slot F1 Score.

\subsubsection{\textsc{Linguist} Results on SNIPS (LNO)} % (fold)
\label{ssub:linguist_for_snips_lno}

We report on the Label Names Only (LNO) setting described in
Section \ref{sec:nifs_lno}.
For these results, we used \textsc{Linguist} models trained without label name dropout,
which we found to perform significantly better (ablation shown in Appendix \ref{sec:ablation_label_names_dropout}).

As show in Table \ref{tab:snips_lno_ic_new} for IC and Table \ref{tab:snips_lno_ner_new} for ST,
despite having \textbf{zero real examples for the new intents,
\textsc{Linguist} achieves 80.0 on Local Intent Recall and 56.9 on Local ST F1 Score}.
(Global metrics are shown in Appendix \ref{sec:results_on_global_metrics_nifs_lno}.)
While this is still far behind using the real text and annotation from these 10 examples
(``s10''), it represents significant progress towards true zero-shot
development of new intents and slots in IC+ST systems.

\begin{table}[h!]
	\footnotesize
\centering
\small
\begin{tabular}{l | c | c }
\hline
 Modified Intent / Data   &   s10      &   \thead{\textsc{Linguist}\\(via s10 LNO)}   \\
\hline
 AddToPlaylist            &  98.3 $\pm$ 2.1 &           68.6 $\pm$ 18.7          \\
 BookRestaurant           &  93.5 $\pm$ 2.4 &           92.1 $\pm$ 5.3           \\
 GetWeather               &  98.8 $\pm$ 0.4 &           99.6 $\pm$ 0.5           \\
 PlayMusic                &   77.1 $\pm$ 6.6 &           85.4 $\pm$ 3.1           \\
 RateBook                 &  99.6 $\pm$ 0.5 &           100.0 $\pm$ 0.0              \\
 SearchCreativeWork       &  76.9 $\pm$ 9.3 &           66.7 $\pm$ 5.8           \\
 SearchScreeningEvent     &   73.1 $\pm$ 8.2 &           47.7 $\pm$ 9.0             \\
 \hline
 Average                  &   88.2 $\pm$ 1.9 &           \textbf{80.0} $\pm$ 2.6             \\
\hline
\end{tabular}
\caption{Local Intent Recall results on SNIPS in the New Intent Few-Shot Label Names Only (NIFS-LNO) setting.
``s10'' results are copied from Table \ref{tab:snips-results-new-ic}.}
\label{tab:snips_lno_ic_new}
\end{table}

\begin{table}[h!]
	\footnotesize
\centering
\small
\begin{tabular}{l | c | c }
\hline
 Modified Intent / Data   &   s10      &   \thead{\textsc{Linguist}\\(via s10 LNO)}   \\
\hline
 AddToPlaylist            & 81.2 $\pm$ 2.5  & 45.9 $\pm$ 9.1          \\
 BookRestaurant           & 81.3 $\pm$ 2.1  & 74.8 $\pm$ 3.5          \\
 GetWeather               & 84.9 $\pm$ 5.4  & 71.2 $\pm$ 3.8          \\
 PlayMusic                & 59.2 $\pm$ 2.2  & 55.6 $\pm$ 3.1          \\
 RateBook                 & 95.0 $\pm$ 0.9    & 55.0 $\pm$ 6.3            \\
 SearchCreativeWork       & 70.9 $\pm$ 11.3 & 59.3 $\pm$ 5.7          \\
 SearchScreeningEvent     & 71.6 $\pm$ 3.6  & 36.6 $\pm$ 6.1          \\
 \hline
 Average                  & 77.7 $\pm$ 2.0    & \textbf{56.9} $\pm$ 2.5          \\
\hline
\end{tabular}
\caption{Local ST F1 Score results on SNIPS in the New Intent Few-Shot Label Names Only (NIFS-LNO) setting.
``s10'' results are copied from Table \ref{tab:snips-results-new-ner}.}
\label{tab:snips_lno_ner_new}
\end{table}

\subsection{mATIS++ Results} % (fold)
\label{sub:matispp_results}

Our mATIS++ results are shown in Tables \ref{tab:matispp_intent} (Intent Accuracy),
and \ref{tab:matispp_slots} (Slot F1).
The main focus is \textbf{``avg-0S'', the average zero-shot performance across the 6 non-en languages} (de, es, fr, hi, ja, pt).

\subsubsection{Baseline Results on mATIS++} % (fold)
\label{ssub:cross_lingual_baseline_results}

An upper bound for zero-shot cross-lingual IC+ST
is multilingual training, where a model is trained jointly on the real data for all languages,
(``all'') which achieves 97.17 for IC and 90.72 for ST.
Reducing to English only data (``en'') harms average zero-shot Intent by 5.0 points,
and Slot F1 by 23.6 points.
As our baseline, we report the numbers from the best cross-lingual 
system (``{MT+soft-align}'') in  \cite{xu20matis},
which uses a specialized transformer architecture for slot alignments,
achieving 94.88 on IC, and 79.84 on ST.

% subsubsection cross_lingual_baseline_results (end).

\subsubsection{Fine-tuning \textsc{Linguist} on MASSIVE} % (fold)
\label{ssub:fine_tuning_on_massive}

% subsubsection fine_tuning_on_massive (end)

We first fine-tune a \textsc{Linguist} model on the MASSIVE dataset, following the
process from Section \ref{sub:training_the_linguist_model}.
We formulate monolingual prompts for each of the 7 languages,
and \emph{cross-lingual} prompts from English to the other 6 languages,
which is straightforward:
for each training utterance in e.g. French, we select up to 10 English training examples that
have the same intent, to include in the prompt with the French utterance as the target,
setting ``French'' in the \texttt{<language>} block of the prompt.

To demonstrate not only cross-lingual and cross-schema
(mATIS++ label names and annotations conventions are different from MASSIVE)
but also
\emph{cross-domain} transfer of \textsc{Linguist}, we exclude the \texttt{transport}
domain from MASSIVE, as it has some overlap with the travel information domain of mATIS++.
We also exclude two other domains for validation early stopping (Appendix \ref{sec:linguits_training_details}).

\subsubsection{\textsc{Linguist} Results on mATIS++} % (fold)
\label{ssub:linguist_results_on_matis}

% subsubsection cross_lingual_generation_on_matis (end)

Then, for inference on mATIS++, we first create monolingual English prompts,
then use a cloud-based MT system to translate the slot values into the target language,
set the \texttt{<language>} tag in the prompt, and generate 10 annotated utterances.
See Figure \ref{fig:flight_fr_1} (Appendix \ref{ssub:cross_lingual_novel_intent_and_slots}) for an example.

We select the output with lowest perplexity,
and use the English IC+ST model to verify its intent,
discarding it if the prediction mismatches the intent from the prompt, in which case we simply
copy over an English utterance from the same intent, to maintain the class distribution.
(See Appendix \ref{sec:ablation_study_on_matis_filtering}.)
The final dataset contains $N$ original English examples,
and $N$ examples for each other language.

Compared to the ``MT+soft-align'' method of \citet{xu20matis},
\textsc{Linguist} is on-par for IC (from 94.88 to 95.06),
and \textbf{improves ST F1 Score by 4.14 points absolute} (from 79.84 to 83.98).
We note that ST, being a structured prediction task, is inherently more challenging
than IC, so our ST results are of particular interest.
Moreover, the improvement on both IC and ST is particularly large for Japanese,
which tends to be challenging for alignment with English,
since the two languages having very different linguistic characteristics.

\begin{table}[h]
	\footnotesize
\centering
\small
\begin{tabular}{c | c | cc || c}
\hline
Lang   &   all &    en & \thead{en+MT\\soft-align}   &   \thead{en+\\\textsc{Linguist}} \\
\hline
 en           & 98.10 & 97.77 & --                 &         97.77 \\
\hline
 de           & 97.32 & 90.51 & 96.66              &         94.08 \\
 es           & 97.21 & 95.20 & 97.20              &         97.10 \\
 fr           & 98.10 & 93.64 & 97.49              &         96.88 \\
 hi           & 95.20 & 88.62 & 92.81              &         94.08 \\
 ja           & 97.86 & 90.99 & 88.33              &         95.38 \\
 pt           & 97.32 & 93.97 & 96.78              &         92.86 \\
\hline
 avg-0S       & 97.17 & 92.16 & 94.88              &         \textbf{95.06} \\
\hline
\end{tabular}
\caption{Results on mATIS++ Intent Accuracy.}
\label{tab:matispp_intent}
\end{table}

\begin{table}[h]
	\footnotesize
\centering
\small
\begin{tabular}{c | c | cc || c}
\hline
Lang   &   all &    en & \thead{en+MT\\soft-align}   &   \thead{en+\\\textsc{Linguist}} \\
\hline
 en           & 95.26 & 95.96 & --                 &         95.07 \\
\hline
 de           & 94.54 & 80.15 & 89.00              &         84.61 \\
 es           & 88.27 & 81.24 & 76.42              &         86.89 \\
 fr           & 92.69 & 77.29 & 79.64              &         83.83 \\
 hi           & 85.58 & 62.61 & 78.56              &         76.61 \\
 ja           & 92.76 & 24.52 & 79.10              &         86.32 \\
 pt           & 90.49 & 76.64 & 76.30              &         85.63 \\
\hline
 avg-0S       & 90.72 & 67.08 & 79.84              &         \textbf{83.98} \\
\hline
\end{tabular}
\caption{Results on mATIS++ Slot F1.}
\label{tab:matispp_slots}
\end{table}

% subsection matis_results (end)

\subsection{Internal Dataset Results}

\label{ssub:internal_dataset_results}
Table \ref{tab:internal-results} shows SemER on our Internal Dataset.

\subsubsection{Baseline Results on Internal Dataset} % (fold)
\label{ssub:baseline_results_for_internal_dataset}

% subsubsection baseline_results_for_internal_dataset (end)
For each feature $i$ we train
an IC+ST model $M_i$
combining the Existing Features data $E$,
up-sampled starter utterances $S_i$,
augmented utterances $A_i$
produced from $S_i$ via Slot Catalog Resampling, ICLM, and BT-Small.
We evaluate on the feature's test set $T_i$, reporting Local SemER.

\subsubsection{\textsc{Linguist} Results on Internal Dataset}
\label{ssub:linguist_internal}

We fine-tune a single \textsc{Linguist} model on instruction prompts formatted from the Existing Features dataset $E$
(Section \ref{ssub:internal_dataset}.),
which \emph{does not contain any examples of the new features}. 
Then, for each feature, $i$, following a similar procedure described in Section \ref{ssub:linguist_for_snips},
we format prompts from the starter utterances $S_i$ and apply \textsc{Linguist} to generate more data $G_i$.
Finally, we follow the same data mixing, training, and testing procedure for each feature $i$ as in the
baseline (Section \ref{ssub:baseline_results_for_internal_dataset}).
As shown in Table \ref{tab:internal-results}, \textsc{Linguist} results in 7.9\% to 25.2\% relative SemER reduction
across four languages compared to the baseline of combined Catalog Resampling, ICLM, and BT-Small.

\begin{table}[h]
\centering
\small
\begin{tabular}{lcccc}
\hline
 \textbf{Feature/Lang}  & \textbf{de}      & \textbf{es}      & \textbf{fr}      & \textbf{ja}      \\
\hline
 CameraControl        & -       & -33.3\% & -       & -       \\
 ClockSettings        & -1.6\%  & -12.3\% & +1.8\%   & -       \\
 HomeSecurity         & -       & -       & -       & -31.5\% \\
 Music                & -36.8\% & -       & -30.6\% & -12.3\% \\
 Timers               & -27.5\% & -15.4\% & -20.0\% & -       \\
\hline
Average                 & \textbf{-11.8\%} & \textbf{-20.8\%} & \textbf{-7.9\%}  & \textbf{-25.2\%} \\
\hline
\end{tabular}
\caption{Internal Dataset Results. Each number is relative reduction in SemER, from baseline (combined Slot Catalog Resampling, BT-Small, and ICLM) to \textsc{Linguist}. A negative number indicates improvement.}
\label{tab:internal-results}
\end{table}

\vspace{-0.2cm}

\section{Conclusion and Future Work}
We introduced \textsc{Linguist}, a novel method for annotated data generation,
via fine-tuning a large-scale pre-trained multilingual seq2seq model.
Our method generalizes to new intents and slots in challenging few-shot, zero-shot, and cross-lingual settings,
which we have shown on three datasets.

In future work, we wish to explore ways to improve the generation output, e.g.
human-in-the-loop and reinforcement learning.
We would also like to include more controls in the prompt such as
text style, and explore generation for multi-turn dialogues, and more complex and nested semantics.

\nopagebreak
\section*{Acknowledgements}

We thank Christophe Dupuy, Weitong Ruan,
and the anonymous reviewers for feedback on our work.

% Entries for the entire Anthology, followed by custom entries
\newpage
\bibliography{alexatm_data_gen_coling_22}
\onecolumn
\appendix
\pagenumbering{arabic}
\renewcommand*{\thepage}{A\arabic{page}} 
\setcounter{figure}{0}    
\renewcommand*{\thefigure}{A\arabic{figure}}

\clearpage
\newpage

\section{Author Contributions} % (fold)
\label{sec:contributions}

Andy proposed and implemented \textsc{Linguist}; performed the experiments on SNIPS, MASSIVE, and mATIS++; fine-tuned \textsc{Linguist} for the internal dataset;
and pre-trained the Alexa Teacher Model 2.3B encoder to warm-start AlexaTM 5B seq2seq.

Saleh implemented the seq2seq pre-training
and fine-tuning code, pre-trained AlexaTM 5B,
provided guidance on experiments and modeling
choices for \textsc{Linguist}, and helped write the paper.

Wael drove the strategic vision for the project,
provided deep technical feedback throughout, and
implemented the backbone of our codebase.

Yannick and Markus assisted with the internal dataset preparation and evaluation,
the baseline methods ICLM
and BT-Small for SNIPS,
and contributed to the paper write-up and technical review.
Yannick also suggested MASSIVE as the dataset for
transferring \textsc{Linguist} to mATIS++.

\section{Sample Model Outputs} % (fold)
\label{sec:sample_model_outputs}

\subsection{English Outputs} % (fold)
\label{sub:english_outputs}

This section uses a \textsc{Linguist} model from 6 SNIPS intents: \texttt{AddToPlaylist}, \texttt{BookRestaurant}, \texttt{PlayMusic}, \texttt{RateBook}, \texttt{SearchCreativeWork}, and \texttt{SearchScreeningEvent}.

% subsection english_outputs (end)

\subsubsection{One-shot Novel Ambiguous Intent} % (fold)
\label{ssub:one_shot_new_intent}

We show (Figure \ref{fig:output_zelda}) the \textsc{Linguist} model generating annotated utterances for a new intent not seen in training data,
given only a single example.
Furthermore, when the example is ambiguous or under-specified,
the model can leverage the intent label
to generate more relevant outputs.

\begin{figure}[h!]
\centering
\begin{subfigure}{.42\textwidth}
  \centering
\begin{Verbatim}[fontsize=\scriptsize, frame=single, commandchars=\\\{\}]
\textbf{INPUT:}
\textcolor{red}{<language>} \textbf{\textcolor{teal}{English}} \textcolor{red}{</language>}
\textcolor{red}{<intent>} \colorbox{yellow!30}{\textbf{\textcolor{violet}{DownloadGame}}} \textcolor{red}{</intent>}
\textcolor{red}{<include>} [1 zelda ] \textcolor{red}{</include>}
\textcolor{red}{<labels>} [1=game \textcolor{red}{</labels>}
\textcolor{red}{<examples>} \textbf{get} [1 zelda ] \textcolor{red}{</examples>}

\textbf{OUTPUTS:}

1. I'd like to play [1 zelda ]
2. I want to play [1 zelda ]
3. I want to \textbf{download} [1 zelda ]
4. I need to \textbf{download} [1 zelda ]
5. Where can I \textbf{download} [1 zelda ] ?
\end{Verbatim}
\vspace{-0.2cm}
\caption{
The prompt with \texttt{DownloadGame} intent.
}
  \label{fig:sub1}
\end{subfigure}%
\hspace{0.05 \textwidth}
\begin{subfigure}{.42\textwidth}
  \centering
\begin{Verbatim}[fontsize=\scriptsize, frame=single, commandchars=\\\{\}]
\textbf{INPUT:}
\textcolor{red}{<language>} \textbf{\textcolor{teal}{English}} \textcolor{red}{</language>}
\textcolor{red}{<intent>} \colorbox{yellow!30}{\textbf{\textcolor{violet}{PurchaseGame}}} \textcolor{red}{</intent>}
\textcolor{red}{<include>} [1 zelda ] \textcolor{red}{</include>}
\textcolor{red}{<labels>} [1=game \textcolor{red}{</labels>}
\textcolor{red}{<examples>} \textbf{get} [1 zelda ] \textcolor{red}{</examples>}

\textbf{OUTPUTS:}

1. I'd like to \textbf{buy} the game [1 zelda ]
2. I'd like to \textbf{buy} [1 zelda ]
3. I want to \textbf{buy} [1 zelda ]
4. Where can I \textbf{buy} [1 zelda ] ?
5. Please \textbf{buy} [1 zelda ]
\end{Verbatim}
\vspace{-0.2cm}
\caption{
The prompt with \texttt{PurchaseGame} intent.
}
  \label{fig:sub2}
\end{subfigure}
\vspace{-0.2cm}
\caption{
For the same example and slots, \textsc{Linguist} can leverage the intent name to generate different outputs.
}
\label{fig:output_zelda}
\end{figure}
\vspace{-0.4cm}
% subsection one_shot_new_intent (end)

\subsubsection{One-shot Novel Ambiguous Slot} % (fold)
\label{sub:one_shot_new_slot}

Similarly, (Figure \ref{fig:output_bat}) from a single example for a novel slot, \textsc{Linguist} can generate
novel values, following the label name to disambiguate an ambiguous word such as ``bat''
(animal or sports equipment).

% Subfigure!!!
\begin{figure}[h!]
\centering
\begin{subfigure}{.48\textwidth}
  \centering
  % \hfill
\begin{Verbatim}[fontsize=\scriptsize, frame=single, commandchars=\\\{\}]
\textbf{INPUT:}
\textcolor{red}{<language>} \textbf{\textcolor{teal}{English}} \textcolor{red}{</language>}
\textcolor{red}{<intent>} \textbf{\textcolor{violet}{DisplayImage}} \textcolor{red}{</intent>}
\textcolor{red}{<include>} [1 \textbf{*} ] \textcolor{red}{</include>}
\textcolor{red}{<labels>} \colorbox{yellow!50}{[1=\textbf{animal}} \textcolor{red}{</labels>}
\textcolor{red}{<examples>}
 show me a picture of a [1 \textbf{bat} ] \textcolor{red}{</examples>}

\textbf{OUTPUTS:}

1. I'd like to see a picture of a [1 \textbf{dog} ]
2. Please show me a picture of a [1 \textbf{deer} ] .
3. Show me a photo of a [1 \textbf{horse} ] .
4. Show me a picture of a [1 \textbf{zebra} ]
5. Show me a [1 \textbf{elephant} ] picture .
\end{Verbatim}
\vspace{-0.2cm}
\caption{
The prompt with \texttt{animal} slot.
}
  \label{fig:sub1}
\end{subfigure}%
% \hspace{0.05 \textwidth}
\hfill
\begin{subfigure}{.48\textwidth}
  \centering
\begin{Verbatim}[fontsize=\scriptsize, frame=single, commandchars=\\\{\}]
\textbf{INPUT:}
\textcolor{red}{<language>} \textbf{\textcolor{teal}{English}} \textcolor{red}{</language>}
\textcolor{red}{<intent>} \textbf{\textcolor{violet}{DisplayImage}} \textcolor{red}{</intent>}
\textcolor{red}{<include>} [1 \textbf{*} ] \textcolor{red}{</include>}
\textcolor{red}{<labels>} \colorbox{yellow!50}{[1=\textbf{sports_equipment}} \textcolor{red}{</labels>}
\textcolor{red}{<examples>}
 show me a picture of a [1 \textbf{bat} ] \textcolor{red}{</examples>}

\textbf{OUTPUTS:}

1. I'd like to see a picture of a [1 \textbf{volleyball} ]
2. Show me a photo of a [1 \textbf{soccer ball} ]
3. Show me a [1 \textbf{football net} ]
4. show me a [1 \textbf{puck} ]
5. Show me a [1 \textbf{golf club} ]
\end{Verbatim}
\vspace{-0.2cm}
\caption{
The prompt with \texttt{sports\_equipment} slot.
}
  \label{fig:sub2}
\end{subfigure}
\vspace{-0.2cm}
\caption{
For the same example and slots, \textsc{Linguist} can leverage the slot name to generate different outputs.
}
\label{fig:output_bat}
\end{figure}

\clearpage
\newpage

\subsubsection{Zero-shot Novel Intent and Slot} % (fold)
\label{sub:zero_shot_new_intent}

We show (Figure \ref{fig:output_hockey}) that \textsc{Linguist} can also generate in a \textbf{zero-shot} setting, prompted only on the intent and slot names,
without any full text examples. By changing the slot type from \texttt{baseball\_team\_name} to
\texttt{hockey\_team\_name}, the model can generate for the requested sports team.
Remarkbly, although the model never saw utterances about sports during fine-tuning, it can accuratey
sample correct team names, presumably relying on the world knowledge it acquired during pre-training.
In the outputs, all baseball teams are actual MLB, and the hockey team names are NHL,
with one exception of \texttt{the Wildcats} being basketball.
The slot \texttt{year} appeared in the SNIPS fine-tuning data, so the model may be
using values either from world knowledge or from other SNIPS intents.

\begin{figure}[h!]
\centering
\begin{subfigure}{.49\textwidth}
  \centering
  % \hfill
\begin{Verbatim}[fontsize=\scriptsize, frame=single, commandchars=\\\{\}]
\textbf{INPUT:}
\textcolor{red}{<language>} \textbf{\textcolor{teal}{English}} \textcolor{red}{</language>}
\textcolor{red}{<intent>} \textbf{\textcolor{violet}{DraftPlayer}} \textcolor{red}{</intent>}
\textcolor{red}{<include>}
 [1 \textbf{*} ] , [2 Patrice Bergeron ] , [3 * ]
\textcolor{red}{</include>}
\textcolor{red}{<labels>}
 \colorbox{yellow!50}{[1=\textbf{baseball_team_name}} , [2=player, [3=year
\textcolor{red}{</labels>}
\textcolor{red}{<examples> </examples>}

\textbf{OUTPUTS:}

1. I want to draft [2 Patrice Bergeron ] from 
   [1 \textbf{Houston Astros} ] for the [3 20th season ].

2. I want to draft [2 Patrice Bergeron ] for the
   [1 \textbf{Washington Nationals} ] in [3 2008 ].
   
3. I'd like to draft [2 Patrice Bergeron ] from 
   [1 \textbf{the Boston Red Sox} ] in
   [3 2019 ]

4. I want to draft [2 Patrice Bergeron ] for 
   [1 \textbf{San Francisco Giants} ] in [3 2007 ]

5. I need to draft [2 Patrice Bergeron ] from 
   [1 \textbf{Toronto Blue Jays} ] for [3 2016 ]
\end{Verbatim}
\caption{
The prompt with \texttt{baseball\_team\_name} intent.
}
  \label{fig:sub1}
\end{subfigure}%
% \hspace{0.05 \textwidth}
\hfill
\begin{subfigure}{.49\textwidth}
  \centering
\begin{Verbatim}[fontsize=\scriptsize, frame=single, commandchars=\\\{\}]
\textbf{INPUT:}
\textcolor{red}{<language>} \textbf{\textcolor{teal}{English}} \textcolor{red}{</language>}
\textcolor{red}{<intent>} \textbf{\textcolor{violet}{DraftPlayer}} \textcolor{red}{</intent>}
\textcolor{red}{<include>}
 [1 \textbf{*} ] , [2 Patrice Bergeron ] , [3 * ]
\textcolor{red}{</include>}
\textcolor{red}{<labels>}
 \colorbox{yellow!50}{[1=\textbf{hockey_team_name}} , [2=player, [3=year
\textcolor{red}{</labels>}
\textcolor{red}{<examples> </examples>}

\textbf{OUTPUTS:}

1. I would like to draft [2 Patrice Bergeron ] from 
   [1 \textbf{Vancouver Canucks} ] in [3 2015 ]

2. I want to draft [2 Patrice Bergeron ] from 
   [1 \textbf{New Jersey Devils} ] in [3 2023 ]

3. Draft the [2 Patrice Bergeron ] from 
   [1 \textbf{the St. Louis Blues} ] for the upcoming 
   [3 2013 ] season.

4. I want to add [2 Patrice Bergeron ] to 
   [1 \textbf{the Wildcats} ] roster for [3 2012 ].

5. Draft [2 Patrice Bergeron ] for 
   [1 \textbf{Boston Bruins} ] in [3 2014 ]
\end{Verbatim}
\caption{
The prompt with \texttt{hockey\_team\_name} intent.
}
  \label{fig:sub2}
\end{subfigure}
\caption{
\textsc{Linguist} can perform zero-shot generation based only on the intent and slot names,
with no full text example.
It can also leverage world knowledge from pre-training to generate
novel slot values that incorporate world knowledge such as valid team names for professional sports.
}
\label{fig:output_hockey}
\end{figure}

\subsubsection{Label Names Only}
\label{sec:example_lno}

Finally, as shown in Figure \ref{fig:output_lno}, the most ambiguous
generation scenario we explore is
``Label Names Only'' (introduced in Section \ref{sec:nifs_lno}), where the model sees no examples,
and all slot values are open-ended via the wildcard instruction \texttt{*}.
Remarkably, \textsc{Linguist} can still generate useful outputs in this case,
although with some more noise: for example, the text of Output 2
does not correspond to GetWeather intent, and the word \texttt{time}
is mistakenly tagged as a slot value \texttt{timeRange}.
Nevertheless, as shown in Section \ref{ssub:linguist_for_snips_lno},
training with \textsc{Linguist}-generated data in this setting can
achieve reasonable accuracy,
representing significant progress towards true zero-shot
data generation for novel intents and slots in IC+ST systems.

\begin{figure}[h!]
\begin{Verbatim}[fontsize=\small, frame=single, commandchars=\\\{\}]
\textbf{INPUT:}
\textcolor{red}{<language>} \textcolor{teal}{\textbf{English}} \textcolor{red}{</language>}
\textcolor{red}{<intent>} \textcolor{violet}{\textbf{GetWeather}} \textcolor{red}{</intent>}
\textcolor{red}{<include>}
 \colorbox{yellow!50}{[1 * ]} , \colorbox{cyan!20}{[2 * ]} , \colorbox{green!20}{[3 * ]}
\textcolor{red}{</include>}
\textcolor{red}{<labels>}
 \colorbox{yellow!50}{[1=city} , \colorbox{cyan!20}{[2=timeRange} , \colorbox{green!20}{[3=state}
\textcolor{red}{</labels>}
\textcolor{red}{<examples> </examples>}

\textbf{OUTPUTS}:

1. What is the weather like for \colorbox{yellow!50}{[1 Oyster Bay ]} \colorbox{green!20}{[3 AK ]} at \colorbox{cyan!20}{[2 midnight ]}

2. What is the \colorbox{red!20}{[2 time ]} for \colorbox{yellow!50}{[1 Red Bank ]}, \colorbox{green!20}{[3 West Virginia ]}

3. What is the weather like in \colorbox{yellow!50}{[1 North Lake Tahoe ]} \colorbox{green!20}{[3 North Dakota ]}
   at \colorbox{cyan!20}{[2 1 minute from now ]}
\end{Verbatim}
\caption{
\textsc{Linguist} generation in Label Names Only (LNO) setting,
where there are no examples or slot values, and the model
must rely entirely on the label names to generate outputs.
}
\label{fig:output_lno}
\end{figure}

% \clearpage
% \newpage

\subsection{Cross-Lingual Novel Intent and Slots} % (fold)
\label{ssub:cross_lingual_novel_intent_and_slots}

In Figure \ref{fig:flight_fr_1} we show an example of \textsc{Linguist} performing \textbf{few-shot cross-lingual annotated data generation on a novel domain},
where it sees only 7 examples in English, and can generate
diverse, fluent, and correctly annotated outputs in French.

The model here was trained on MASSIVE both monolingual and cross-lingual prompts, as described in Section \ref{ssub:fine_tuning_on_massive}.
We discarded three of the 18 MASSIVE domains during training, namely \texttt{audio} and \texttt{cooking}
to use as validation sets, and \texttt{transport}, in order to keep mATIS++, which covers travel information, 
as a novel domain.

\begin{figure}[h!]
\begin{Verbatim}[fontsize=\small, frame=single, commandchars=\\\{\}]
\textbf{INPUT:}
\textcolor{red}{<language>} \colorbox{orange!20}{\textbf{\textcolor{teal}{French}}} \textcolor{red}{</language>}
\textcolor{red}{<domain>} \textbf{\textcolor{blue}{travelinfo}} \textcolor{red}{</domain>}
\textcolor{red}{<intent>} \textbf{\textcolor{violet}{flight}} \textcolor{red}{</intent>}
\textcolor{red}{<include>} \colorbox{yellow!50}{[2 lundi ]} , \colorbox{cyan!20}{[3 baltimore ]} , \colorbox{green!20}{[4 dallas ]} \textcolor{red}{</include>}
\textcolor{red}{<labels>}
 [1=depart_time.period_of_day , \colorbox{yellow!50}{[2=depart_date.day_name} , 
 \colorbox{cyan!20}{[3=toloc.city_name} , \colorbox{green!20}{[4=fromloc.city_name}
\textcolor{red}{</labels>}
\textcolor{red}{<examples>}
     show me the flights from [4 atlanta ] to [3 denver ] on [2 friday ]
\textcolor{red}{<br>} what flights go from [4 charlotte ] to [3 baltimore ] in the [1 morning ]
\textcolor{red}{<br>} show me all the flights from [4 philadelphia ] to [3 cincinnati ]
\textcolor{red}{<br>} flights on [2 friday ] from [4 newark ] to [3 tampa ]
\textcolor{red}{<br>} what flights leave from [4 chicago ] to [3 seattle ] on [2 saturday ]
\textcolor{red}{<br>} list flights from [4 detroit ] to [3 san diego ] on [2 thursday ]
\textcolor{red}{<br>} [4 dallas ] to [3 baltimore ] [2 monday ]
\textcolor{red}{</examples>}

\textbf{OUTPUTS:}

1. combien de vols y a-t-il entre \colorbox{green!20}{[4 dallas ]} et \colorbox{cyan!20}{[3 baltimore ]} le \colorbox{yellow!50}{[2 lundi ]}

2. quels vols y a-t-il \colorbox{yellow!50}{[2 lundi ]} entre \colorbox{green!20}{[4 dallas ]} et \colorbox{cyan!20}{[3 baltimore ]}

3. quel est le meilleur vol pour \colorbox{yellow!50}{[2 lundi ]} entre \colorbox{green!20}{[4 dallas ]} et \colorbox{cyan!20}{[3 baltimore ]}

4. s'il vous plait trouvez les vols partant de \colorbox{green!20}{[4 dallas ]} le \colorbox{yellow!50}{[2 lundi ]}
   à destination de \colorbox{cyan!20}{[3 baltimore ]}

5. quelles sont les vols qui partent le \colorbox{yellow!50}{[2 lundi ]} de \colorbox{green!20}{[4 dallas ]}
   à \colorbox{cyan!20}{[3 baltimore ]}
\end{Verbatim}
	\caption{\textsc{Linguist} generating on a Cross-lingual English to French example from mATIS++.}
\label{fig:flight_fr_1}
\end{figure}

\section{\textsc{Linguist} Training Details} % (fold)
\label{sec:linguits_training_details}

\subsection{Hyperparamters and Early Stopping}
\label{sec:early_stopping}

\begin{figure}[h!]
\centering
\begin{subfigure}{.49\textwidth}
  \centering
  \includegraphics[width=0.9\textwidth]{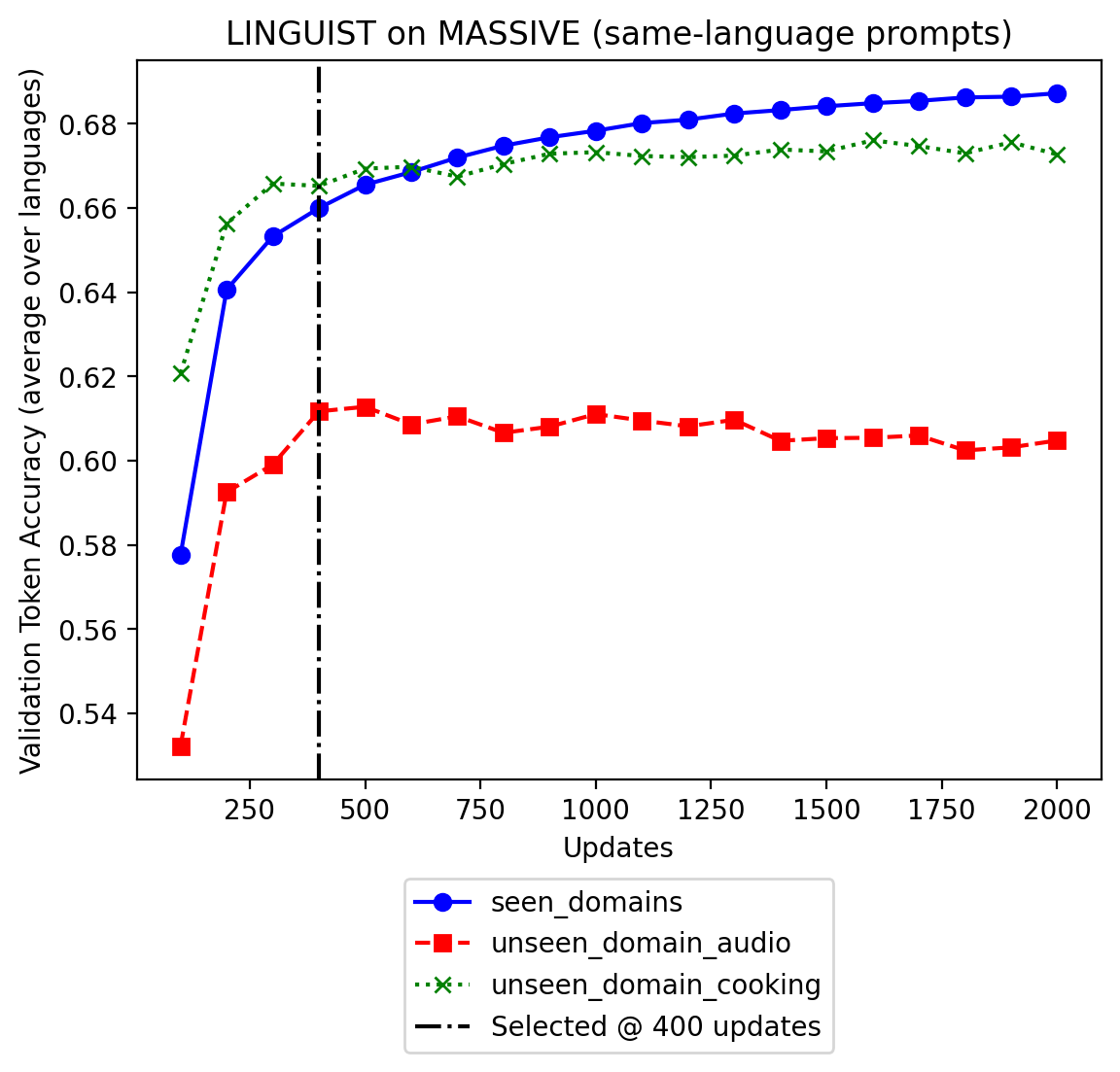}

\caption{
Validation Token Accuracy on same-language prompts.
}
  \label{fig:sub1}
\end{subfigure}%
% \hspace{0.05 \textwidth}
\hfill
\begin{subfigure}{.49\textwidth}
  \centering
\includegraphics[width=0.9\textwidth]{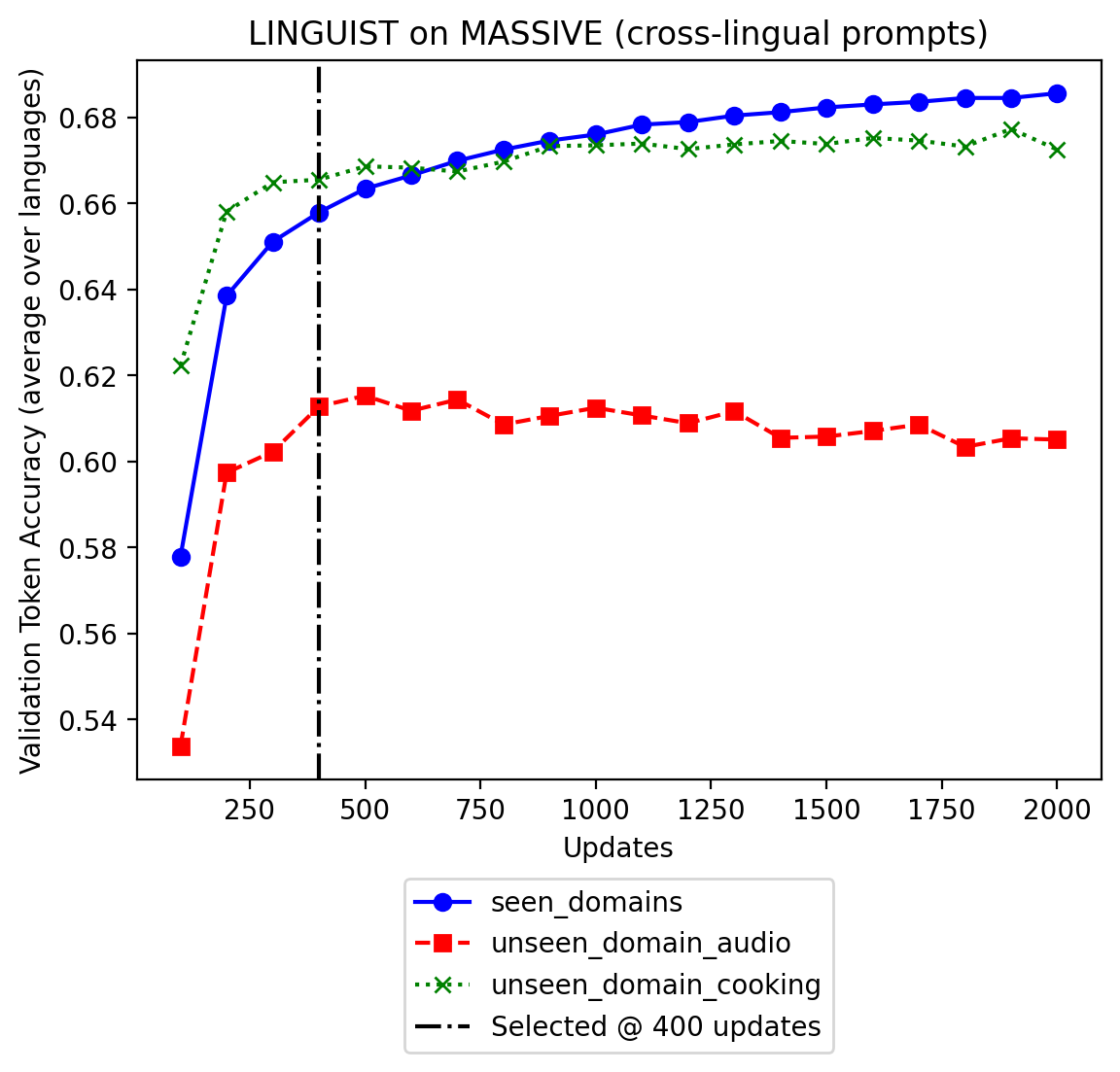}
\caption{
Validation Token Accuracy on cross-lingual prompts.
}
  \label{fig:sub2}
\end{subfigure}
\caption{
Validation Accuracy across updates of fine-tuning \textsc{Linguist} on the MASSIVE dataset.
}
\label{fig:val_tok_acc}
\end{figure}

As shown in Figure \ref{fig:val_tok_acc}, we find that the model converges after a very small number of updates.
Specifically, we train with batch size 512 for 400 updates (i.e. around 18 epochs for SNIPS, around 2.5 epochs for MASSIVE),
using a very small learning rate 5e-7 with Adam \citep{Kingma2015AdamAM}, warmed up over the first 100 updates,
then kept constant for the rest of training.
(The internal dataset is much larger, so for that, we train for 4k updates instead, and use a larger learning rate of 1e-6.)

For MASSIVE, we removed two additional small domains \texttt{audio} and \texttt{cooking} from training
and early stop once Token Accuracy plateaus on these domains, which occurs around 400 updates.
The Token Accuracy is the percentage of subword tokens for which the model's top-1 hypothesis
matches the ground truth (higher is better).
It requires only a single forward pass to compute, without needing
auto-regressive decoding.
In early experiments, we found Token Accuracy to be more reliable than perplexity
at predicting downstream performance.

As shown in Figure \ref{fig:val_tok_acc}, the Token Accuracy continues to improve
for the domains seen during training, however plateaus after 400 updates for the two novel domains,
suggesting overfitting beyond 400 updates.
The token accuracy is similar for same-language (left) and cross-lingual (right) prompts,
suggesting that the model can jointly learn both tasks to a similar level of performance.

For the SNIPS runs, the data is so limited to only 6 intents per run,
so removing another intent to check for early stopping could harm the model performance.
Therefore, we simply use 400 updates again, as that worked for MASSIVE.

We use DeepSpeed \citep{Rasley2020DeepSpeed} ZeRO Stage 2 to accelerate training.

% Subfigure!!!

\subsection{Tokenizer Choices}
\label{sec:tokenizer_choices}

As mentioned in Section \ref{sub:training_the_linguist_model},
we keep the original sentencepice \citep{kudo-richardson-2018-sentencepiece}
tokenizer of the model; we do not explicitly add vocabulary
items for \texttt{<intent>}, \texttt{[1}, etc.
This choice is motivated by two intuitions:

(1) We hypothesize that these tags' resemblance to markup languages like HTML/XML
may help the model learn that they are instructions rather than content words,
by relying on data seen during pre-training which was formatted similarly.
An earlier version of the \textsc{Linguist} prompt had only the open tags such as
\texttt{<intent PlayMusic>}, and we found that the pre-trained model
produced matching closing tags such as \texttt{</intent>} before any fine-tuning,
suggesting it had learned some knowledge of the tag structure from pre-training.

(2) We hypothesize that the model may be able to generalize at inference time to
utterances with a larger number of slots than seen during fine-tuning,
by tokenizing the numbered brackets.
For example, the largest numbered bracket seen when fine-tuning on MASSIVE is \texttt{[10},
tokenized as \texttt{['\_[', '10']}.
For inference on mATIS++, 47\% of the prompts contain numbered brackets between
\texttt{[11} and \texttt{[24}.
If we had added vocabulary items for these, they would be stuck as randomly initialized tokens
at inference time, and therefore unlikely to produce high quality generation.
Instead, we see in the generated outputs many examples where the model handles these larger
numbered brackets without a problem.

% section training_details (end)

\section{Impact of Model Size}
\label{sec:model_size}

To evaluate the impact of the model size used for \textsc{Linguist} data generation,
we evaluate on mATIS++ with a 10x smaller model, following the same procedure of first
fine-tuning on MASSIVE (Section \ref{ssub:fine_tuning_on_massive}), then running cross-lingual
inference on mATIS++  (Section \ref{ssub:linguist_results_on_matis}).

We use \textbf{AlexaTM-Large 500M}, which is trained using the same data as AlexaTM 20B \citep{Soltan2022AlexaTM2F}
and AlexaTM 5B (Section \ref{sub:data_gen_mdoel}).
Like AlexaTM 5B, AlexaTM-Large 500M uses only the denoising objective
(no Causal Language Modeling).
The architecture contains 12 encoder and 12 decoder layers, with hidden size 1024,
the same as (m)BART \citep{lewis-etal-2020-bart,liu20mbart}.

As show in Tables \ref{tab:matispp_intent_model_size} for IC and \ref{tab:matispp_slots_model_size} for ST,
switching to the smaller model loses 1.62 points on IC (from 95.06 to 93.44),
and 1.74 points on ST (from 83.98 to 82.24).
While \textsc{Linguist} with the smaller model
under-performs the baseline ``en+MT soft-align'' on IC by 1.44 points (93.44 compared to 94.88),
it still out-performs on ST by 2.40 points (82.24 compared to 79.84),
showing the value of the \textsc{Linguist} on the more challenging ST task,
even with a smaller model.

\begin{table}[h]
	\footnotesize
\centering
\small
\begin{tabular}{c | c | cc || c | c}
\hline
Lang   &   all &    en & \thead{en+MT\\soft-align}   &   \thead{en+\\\textsc{Linguist}} & \thead{en+\\\textsc{Linguist}\\(AlexaTM-Large 500M)} \\
 \hline
 en           & 98.10 & 97.77 & --                 &         97.77 &                97.88 \\
 \hline
 de           & 97.32 & 90.51 & 96.66              &         94.08 &                95.65 \\
 es           & 97.21 & 95.20 & 97.2               &         97.10 &                95.87 \\
 fr           & 98.10 & 93.64 & 97.49              &         96.88 &                95.98 \\
 hi           & 95.20 & 88.62 & 92.81              &         94.08 &                86.94 \\
 ja           & 97.86 & 90.99 & 88.33              &         95.38 &                91.55 \\
 pt           & 97.32 & 93.97 & 96.78              &         92.86 &                94.64 \\
\hline
 avg-0S       & 97.17 & 92.16 & 94.88              &         \textbf{95.06} &                93.44 \\
\hline
\end{tabular}
\caption{Results on mATIS++ Intent Accuracy, showing impact of model size. All except ``en+\textsc{Linguist} (AlexaTM-Large 
500M)'' are copied from Table \ref{tab:matispp_intent}.}
\label{tab:matispp_intent_model_size}
\end{table}
% thead{Train\\ /Test}

\begin{table}[h]
	\footnotesize
\centering
\small
\begin{tabular}{c | c | cc || c | c}
\hline
Lang   &   all &    en & \thead{en+MT\\soft-align}   &   \thead{en+\\\textsc{Linguist}} & \thead{en+\\\textsc{Linguist}\\(AlexaTM-Large 500M)} \\
\hline
 en           & 95.26 & 95.96 & --                 &         95.07 &                95.51 \\
\hline
 de           & 94.54 & 80.15 & 89                 &         84.61 &                83.02 \\
 es           & 88.27 & 81.24 & 76.42              &         86.89 &                85.45 \\
 fr           & 92.69 & 77.29 & 79.64              &         83.83 &                82.19 \\
 hi           & 85.58 & 62.61 & 78.56              &         76.61 &                75.36 \\
 ja           & 92.76 & 24.52 & 79.1               &         86.32 &                83.23 \\
 pt           & 90.49 & 76.64 & 76.3               &         85.63 &                84.18 \\
 \hline
 avg-0S       & 90.72 & 67.08 & 79.84              &         \textbf{83.98} &                82.24 \\
\hline
\end{tabular}
\caption{Results on mATIS++ Slot F1, showing impact of model size. All except ``en+\textsc{Linguist} (AlexaTM-Large 
500M)'' are copied from Table \ref{tab:matispp_slots}.}
\label{tab:matispp_slots_model_size}
\end{table}

\section{Ablation Study on Label Name Dropout}
\label{sec:ablation_label_names_dropout}

In early experiments, we found Label Name Dropout (LNDrop, described in Section \ref{sub:training_the_linguist_model})
helped reduce the model's tendency to overfit on the label
names seen during training.
However, at the end of our experiments,
the ablation study in this section (Table \ref{tab:snips_no_drop_ic_new} for 
Local IC Recall and Table \ref{tab:snips_no_drop_ner_new}
for Local ST F1 Score) shows an improvement from
removing the label dropout.
The impact is small on ``s10'' (where the model sees both the label names and the annotated
text for the 10 starter examples), however is quite larger
in ``\textsc{Linguist} (via s10 LNO)'' (i.e. Label Names Only, Section \ref{sec:nifs_lno}),
where the model sees only the label names.

We hypothesize that other changes we made along the way such as
reducing the number of model updates via early stopping (Section 
\ref{sec:early_stopping}) and reducing the learning rate
may have introduced regularization which makes label name dropout
less necessary, and may surface a side effect of adding undesirable noise.
In future work, we would like to study this more thoroughly,
and investigate whether label dropout helps in cases
where at inference time, the label names are either not available,
or are not descriptive of the utterance semantics.

\begin{table}[h!]
	\footnotesize
\centering
\small
\begin{tabular}{l | c | c || c | c }
\hline
 Modified Intent / Data   &   \thead{s10\\+\textsc{Linguist}\\LNDrop: yes}      &   \thead{s10\\+\textsc{Linguist}\\LNDrop: \textbf{no}}  & 
 \thead{\textsc{Linguist}\\(via s10 LNO)\\LNDrop: yes} &
  \thead{\textsc{Linguist}\\(via s10 LNO)\\LNDrop: \textbf{no}} \\
  \hline
 AddToPlaylist            & 93.9 $\pm$ 3.2                  & 92.3 $\pm$ 7.8                 & 20.0 $\pm$ 11.6                      & 68.6 $\pm$ 18.7                     \\
 BookRestaurant           & 94.6 $\pm$ 1.5                  & 93.8 $\pm$ 2.3                 & 86.5 $\pm$ 4.9                       & 92.1 $\pm$ 5.3                      \\
 GetWeather               & 100.0 $\pm$ 0.0                 & 99.6 $\pm$ 0.5                 & 99.2 $\pm$ 0.8                       & 99.6 $\pm$ 0.5                      \\
 PlayMusic                & 90.4 $\pm$ 4.7                  & 90.0 $\pm$ 3.1                 & 76.2 $\pm$ 6.0                       & 85.4 $\pm$ 3.1                      \\
 RateBook                 & 100.0 $\pm$ 0.0                 & 99.8 $\pm$ 0.4                 & 99.6 $\pm$ 0.5                       & 100.0 $\pm$ 0.0                     \\
 SearchCreativeWork       & 83.3 $\pm$ 6.9                  & 92.5 $\pm$ 4.0                 & 66.1 $\pm$ 13.4                      & 66.7 $\pm$ 5.8                      \\
 SearchScreeningEvent     & 81.9 $\pm$ 3.9                  & 79.4 $\pm$ 2.8                 & 44.2 $\pm$ 6.7                       & 47.7 $\pm$ 9.0                      \\
 \hline
 Average                  & 92.0 $\pm$ 0.8                  & \textbf{92.5} $\pm$ 1.5                 & 70.3 $\pm$ 2.6                       & \textbf{80.0} $\pm$ 2.6 \\
\hline
\end{tabular}
\caption{Local Intent Recall results on SNIPS comparing
with and without Label Name Dropout (LNDrop), in settings
NIFS vanilla (left two columns) and NIFS-LNO (Label Names Only, Section \ref{sec:nifs_lno})
(right two columns).
``s10+\textsc{Linguist} (LNDrop: yes)'' results are copied from Table \ref{tab:snips-results-new-ic} and ``\textsc{Linguist} (via s10 LNO) LNDrop: no'' results
are copied from Table \ref{tab:snips_lno_ic_new}.}
\label{tab:snips_no_drop_ic_new}
\end{table}

\begin{table}[h!]
	\footnotesize
\centering
\small
\begin{tabular}{l | c | c || c | c }
\hline
 Modified Intent / Data   &   \thead{s10\\+\textsc{Linguist}\\LNDrop: yes}      &   \thead{s10\\+\textsc{Linguist}\\LNDrop: \textbf{no}}  & 
 \thead{\textsc{Linguist}\\(via s10 LNO)\\LNDrop: yes} &
  \thead{\textsc{Linguist}\\(via s10 LNO)\\LNDrop: \textbf{no}} \\
  \hline
 AddToPlaylist            & 80.9 $\pm$ 3.4                  & 80.4 $\pm$ 2.1                 & 56.6 $\pm$ 8.0                       & 45.9 $\pm$ 9.1                      \\
 BookRestaurant           & 83.4 $\pm$ 1.7                  & 82.8 $\pm$ 4.0                 & 70.7 $\pm$ 3.5                       & 74.8 $\pm$ 3.5                      \\
 GetWeather               & 85.4 $\pm$ 2.8                  & 83.1 $\pm$ 3.6                 & 70.0 $\pm$ 3.4                       & 71.2 $\pm$ 3.8                      \\
 PlayMusic                & 70.1 $\pm$ 1.8                  & 69.3 $\pm$ 2.5                 & 54.2 $\pm$ 3.2                       & 55.6 $\pm$ 3.1                      \\
 RateBook                 & 94.8 $\pm$ 1.7                  & 96.1 $\pm$ 1.0                 & 51.0 $\pm$ 11.1                      & 55.0 $\pm$ 6.3                      \\
 SearchCreativeWork       & 79.3 $\pm$ 5.0                  & 85.3 $\pm$ 3.7                 & 55.5 $\pm$ 11.2                      & 59.3 $\pm$ 5.7                      \\
 SearchScreeningEvent     & 82.3 $\pm$ 3.4                  & 80.4 $\pm$ 1.4                 & 29.7 $\pm$ 2.9                       & 36.6 $\pm$ 6.1                      \\
 \hline
 Average                  & 82.3 $\pm$ 1.3                  & \textbf{82.5} $\pm$ 1.7                 & 55.4 $\pm$ 2.3                       & \textbf{56.9} $\pm$ 2.5                      \\
\hline
\end{tabular}
\caption{Local ST F1 Score results on SNIPS comparing
with and without Label Name Dropout (LNDrop), in settings
NIFS vanilla (left two columns) and NIFS-LNO (Label Names Only, Section \ref{sec:nifs_lno})
(right two columns).
``s10+\textsc{Linguist} (LNDrop: yes)'' results are copied from Table \ref{tab:snips-results-new-ner} and ``\textsc{Linguist} (via s10 LNO) LNDrop: no'' results
are copied from Table \ref{tab:snips_lno_ner_new}.}
\label{tab:snips_no_drop_ner_new}
\end{table}

\section{Ablation Study on Filtering Generated Annotated Utterances for mATIS++} % (fold)
\label{sec:ablation_study_on_matis_filtering}

We present an ablation study on filtering the outputs of \textsc{Linguist} for mATIS++, with results presented in
Table \ref{tab:matispp_intent_abl} for Intent Accuracy, and Table \ref{tab:matispp_slots_abl} for Slot F1.

\subsection{Filtering Methods} % (fold)
\label{sub:filtering_methods}

As introduced in Section \ref{ssub:linguist_results_on_matis}, for each annotated English utterance in mATIS++,
we generate 10 annotated utterances in each of the zero-shot languages (German, Spanish, French, Hindi, Japanese, Portuguese).
Then, for each language, we select the single\footnote{
In future work, we would like to evaluate the impact on mATIS++ of including more than one output per input prompt, 
as we have done for SNIPS and for the Internal Dataset.} 
generated annotated utterance with lowest perplexity which also passes \texttt{Valid-Filter}, described next.

We apply two filtering methods, (1) \texttt{Valid-Filter}, and (2) \texttt{English-IC-Filter}, and report the ``Pass Rate''
of each in Table \ref{tab:matispp_success}, as the portion of utterances that pass the filter.

For \texttt{Valid-Filter} we discard utterances that have invalid brackets like \texttt{[2 [ ]}, or do not respect the
prompt, by either generating too few or too many slots, or not copying the value when requested.
The Pass Rate for \texttt{Valid-Filter} filter is 71.5\%, averaged across the languages.

For \texttt{English-IC-Filter}, we classify the intent of the generated utterance's text, using the English-only IC+ST model,
and discard the utterance if the predicted intent disagrees with the intent from the \textsc{Linguist} prompt.
We note that the English-only IC+ST model (``en'' in Table \ref{tab:matispp_intent_abl}) already performs quite well
on IC for other languages, achieving 92.16 Intent Accuracy, so we expect it to contain a strong signal to filter out noisy generated utterances.
The Pass Rate for \texttt{English-IC-Filter} is 83.8\%, suggesting that the remaining 16.2\% of the utterances
are likely to correspond to an intent other than what was requested in the prompt,
which we discuss further below (Section \ref{sub:intent_mismatch_discussion}).

After cascading the two filters, the overall Pass Rate is 59.9\%.

As a final step, observing that some intents may have lost more data than others, we apply a simple fix which we call
\texttt{Balance-Classes} to recover the original
per-intent class distribution: we simply copy over English utterances from the intents that
lack enough data.

\begin{table}[h!]
	\footnotesize
\centering
\small
\begin{tabular}{c | c c | c | c c c}
\hline
 Lang   &   \thead{\texttt{Valid-}\\\texttt{Filter}\\Pass Rate} &   \thead{\texttt{English-}\\\texttt{IC-Filter}\\Pass Rate} &   \thead{Cascaded\\Pass Rate} &  \thead{Num Outputs\\from \textsc{Linguist}} &   \thead{Num Copied\\from English} &   \thead{Total} \\
\hline
 de     &                 73.8 &                     77.1 &                 56.9 &                   2393 &                         1816 &    4209 \\
 es     &                 77.2 &                     88.2 &                 68.1 &                   2867 &                         1342 &    4209 \\
 fr     &                 77.3 &                     80.2 &                 62.0 &                   2609 &                         1600 &    4209 \\
 hi     &                 75.4 &                     85.6 &                 64.6 &                   2717 &                         1492 &    4209 \\
 ja     &                 50.5 &                     84.8 &                 42.8 &                   1801 &                         2408 &    4209 \\
 pt     &                 74.7 &                     87.2 &                 65.2 &                   2744 &                         1465 &    4209 \\
\hline
 avg    &                 71.5 &                     83.8 &                 59.9 &                   2522 &                         1687 &    4209 \\
\hline
\end{tabular}
\caption{Pass Rate of \textsc{Linguist} generated utterances for mATIS++}
\label{tab:matispp_success}
\end{table}

\subsection{Impact of Filtering} % (fold)
\label{sub:impact_of_filtering}

The impact of filtering is presented in Table \ref{tab:matispp_intent_abl} for Intent Accuracy, and Table \ref{tab:matispp_slots_abl} for Slot F1,
where our main result is ``avg-0S'', the average of the zero-shot languages (de, es, fr, hi, ja, pt).
We observe that \texttt{English-IC-Filter} improves IC by 0.89 points absolute (from 93.09 to 93.98) and \texttt{Balance-Classes}
improves IC by a further 1.08 points absolute (from 93.98 to 95.06), with both methods having minimal impact on Slot F1.

\begin{table}[h!]
	\footnotesize
\centering
\small
\begin{tabular}{c | c | cc || c c c}
\hline
Lang   &   all &    en & \thead{en+MT\\soft-align}   &   \thead{en+\\\textsc{Linguist}\\(NoFilter)} &   \thead{en+\\\textsc{Linguist}\\(+English-IC-Filter)} &   \thead{en+\\\textsc{Linguist}\\(+English-IC-Filter)\\(+Balance-Classes)} \\
\hline
 en           & 98.10 & 97.77 & --                 &                  97.21 &                97.77 &                       97.77 \\
\hline
 de           & 97.32 & 90.51 & 96.66              &                  94.31 &                93.64 &                       94.08 \\
 es           & 97.21 & 95.20 & 97.2               &                  94.75 &                96.88 &                       97.10 \\
 fr           & 98.10 & 93.64 & 97.49              &                  93.97 &                96.43 &                       96.88 \\
 hi           & 95.20 & 88.62 & 92.81              &                  88.17 &                92.19 &                       94.08 \\
 ja           & 97.86 & 90.99 & 88.33              &                  91.78 &                91.44 &                       95.38 \\
 pt           & 97.32 & 93.97 & 96.78              &                  95.54 &                93.30 &                       92.86 \\
\hline
 avg-0S       & 97.17 & 92.16 & 94.88              &                  93.09 &                93.98 &                       \textbf{95.06} \\
\hline
\end{tabular}
\caption{Intent Accuracy results on ablation study of Filtering for mATIS++.}
\label{tab:matispp_intent_abl}
\end{table}

% thead{Train\\ /Test}

\begin{table}[h!]
	\footnotesize
\centering
\small
\begin{tabular}{c | c | cc || c c c}
\hline
Lang   &   all &    en & \thead{en+MT\\soft-align}   &   \thead{en+\\\textsc{Linguist}\\(NoFilter)} &   \thead{en+\\\textsc{Linguist}\\(+English-IC-Filter)} &   \thead{en+\\\textsc{Linguist}\\(+English-IC-Filter)\\(+Balance-Classes)} \\
\hline
 en           & 95.26 & 95.96 & --                 &                  94.16 &                95.01 &                       95.07 \\
\hline
 de           & 94.54 & 80.15 & 89                 &                  83.40 &                85.26 &                       84.61 \\
 es           & 88.27 & 81.24 & 76.42              &                  85.92 &                85.47 &                       86.89 \\
 fr           & 92.69 & 77.29 & 79.64              &                  84.65 &                84.71 &                       83.83 \\
 hi           & 85.58 & 62.61 & 78.56              &                  78.35 &                75.89 &                       76.61 \\
 ja           & 92.76 & 24.52 & 79.1               &                  85.72 &                85.38 &                       86.32 \\
 pt           & 90.49 & 76.64 & 76.3               &                  84.45 &                85.06 &                       85.63 \\
\hline
 avg-0S       & 90.72 & 67.08 & 79.84              &                  83.75 &                83.63 &                       \textbf{83.98} \\
\hline
\end{tabular}
\caption{Slot F1 results on ablation study of Filtering for mATIS++.}
\label{tab:matispp_slots_abl}
\end{table}

\subsection{Intent Mismatch Discussion} % (fold)
\label{sub:intent_mismatch_discussion}

We discuss an intuition about why \textsc{Linguist} in a cross-domain setting such as MASSIVE to mATIS++
might produce outputs that do not exactly match the prompted intent.
Notice that the prompt contains only 10 examples of the \emph{target} intent, and no examples of \emph{other} intents from the
new domain.
For example, mATIS++ contains several closely related intents, such as ``flight'' which asks to list and book flights,
``flight\_time'' which asks about the \emph{time} of a flight, and ``airfare'' which asks about the \emph{price} of a flight.
We find that when the prompts contain only ``flight'' examples, the model tends to over-generalize and produce some
requests asking for time or price instead, which harms IC, necessitating a post-processing method such as \texttt{English-IC-Filter}.
In future work, we plan to explore methods to incorporate few-shot data from \emph{other} intents in the domain
while generating for the target intent, to mitigate this problem.

% section ablation_study_on_matis_filtering (end)

\section{SNIPS Results on Global Metrics} % (fold)
\label{sec:results_on_global_metrics}
We report the resutls on SNIPS Global Metrics as mentioned in Section \ref{sub:snips_exp}.

\subsection{SNIPS Results on Global Metrics: NIFS}
\label{sec:results_on_global_metrics_nifs}

Our main results are on the New-Intent Few-Shot setting (NIFS, described in Section \ref{ssub:new_intent_few_shot_nifs}).
Table \ref{tab:snips-results-overall-ic} shows Global Intent Accuracy, 
and Table \ref{tab:snips-results-overall-ner} shows Global ST F1 Score.
These correspond to the Local metrics shown in Tables \ref{tab:snips-results-new-ic} for Local IC and \ref{tab:snips-results-new-ner} for Local ST.

As all our methods target a new-intent setting, as expected, they do not substantially
impact the Global metrics. Nonetheless,
\textsc{Linguist} does provide a small improvement of +0.3 points absolute on both IC and ST compared to Ex2.

\begin{table*}[h!]
\setlength{\tabcolsep}{4.0pt}
\begin{subtable}[h]{1\textwidth}
\centering
\small
\makebox[\textwidth][c]{
\begin{tabular}{lcc | c | cccc | c}
\hline
 Modified Intent / Data   &   Full & s10-NoUps        & \thead{s10}      & \thead{s10 \\ +ICLM}   & \thead{s10 \\ +BT-Small}   & \thead{s10 \\ +BT-5B}    & \thead {s10 \\ +Ex2}   & \thead{s10 \\ +\textbf{\textsc{Linguist}}}   \\
\hline
 AddToPlaylist            &   99.1 & 98.7 $\pm$1.0 & 98.9 $\pm$0.3 & 98.9 $\pm$0.3   & 98.9 $\pm$0.3       & 99.1 $\pm$0.1 & 99.0 $\pm$0.3  & 98.4 $\pm$0.6       \\
 BookRestaurant           &   99.1 & 97.7 $\pm$0.4 & 98.2 $\pm$0.3 & 98.2 $\pm$0.2   & 97.9 $\pm$0.2       & 98.2 $\pm$0.1 & 97.9 $\pm$0.8  & 98.2 $\pm$0.3       \\
 GetWeather               &   99.1 & 98.9 $\pm$0.1 & 98.9 $\pm$0.1 & 98.9 $\pm$0.2   & 99.0 $\pm$0.0       & 99.0 $\pm$0.1 & 99.1 $\pm$0.2  & 99.1 $\pm$0.1       \\
 PlayMusic                &   99.1 & 95.3 $\pm$1.5 & 96.1 $\pm$1.0 & 96.4 $\pm$1.3   & 96.0 $\pm$0.8       & 97.0 $\pm$0.4 & 97.1 $\pm$1.1  & 97.9 $\pm$0.6       \\
 RateBook                 &   99.1 & 99.0 $\pm$0.1 & 98.9 $\pm$0.1 & 99.0 $\pm$0.1   & 99.0 $\pm$0.1       & 99.0 $\pm$0.1 & 99.0 $\pm$0.0  & 99.0 $\pm$0.1       \\
 SearchCreativeWork       &   99.1 & 95.2 $\pm$1.1 & 96.5 $\pm$1.2 & 96.0 $\pm$1.2   & 95.8 $\pm$1.8       & 96.7 $\pm$1.4 & 96.7 $\pm$0.8  & 97.3 $\pm$1.0       \\
 SearchScreeningEvent     &   99.1 & 95.3 $\pm$0.8 & 96.3 $\pm$1.1 & 95.9 $\pm$0.5   & 95.8 $\pm$0.8       & 96.2 $\pm$1.5 & 96.8 $\pm$1.2  & 97.4 $\pm$0.5       \\
\hline
 Average                  &   99.1 & 97.2 $\pm$0.4 & 97.7 $\pm$0.2 & 97.6 $\pm$0.4   & 97.5 $\pm$0.3       & 97.9 $\pm$0.2 & 97.9 $\pm$0.3  & \textbf{98.2} $\pm$0.1       \\
\hline
\end{tabular}
}
\caption{SNIPS New-Intent few-shot results on \textbf{Global Intent Accuracy}.}
\label{tab:snips-results-overall-ic}
\end{subtable}
\hfill

\begin{subtable}[h]{1\textwidth}
\centering
\small
\makebox[\textwidth][c]{
\begin{tabular}{lcc | c | cccc | c}
\hline
 Modified Intent / Data   &   Full & s10-NoUps        & \thead{s10}      & \thead{s10 \\ +ICLM}   & \thead{s10 \\ +BT-Small}   & \thead{s10 \\ +BT-5B}    & \thead {s10 \\ +Ex2}   & \thead{s10 \\ +\textbf{\textsc{Linguist}}}   \\
\hline
 AddToPlaylist            &   96.7 & 94.0 $\pm$0.4 & 94.2 $\pm$0.5 & 94.2 $\pm$0.4   & 94.9 $\pm$0.4       & 94.5 $\pm$0.3 & 94.7 $\pm$0.4  & 94.6 $\pm$0.6       \\
 BookRestaurant           &   96.7 & 92.7 $\pm$0.4 & 94.1 $\pm$0.6 & 94.1 $\pm$0.3   & 94.2 $\pm$0.3       & 94.5 $\pm$0.3 & 93.9 $\pm$1.0  & 94.6 $\pm$0.4       \\
 GetWeather               &   96.7 & 93.8 $\pm$0.5 & 94.9 $\pm$0.7 & 94.7 $\pm$0.6   & 94.6 $\pm$0.4       & 95.0 $\pm$0.3 & 94.7 $\pm$0.7  & 95.0 $\pm$0.4       \\
 PlayMusic                &   96.7 & 91.3 $\pm$0.4 & 92.9 $\pm$0.3 & 93.1 $\pm$0.6   & 92.8 $\pm$0.2       & 93.8 $\pm$0.4 & 93.9 $\pm$0.6  & 94.2 $\pm$0.2       \\
 RateBook                 &   96.7 & 95.0 $\pm$0.3 & 95.8 $\pm$0.4 & 95.8 $\pm$0.3   & 95.5 $\pm$0.6       & 95.5 $\pm$0.4 & 96.0 $\pm$0.2  & 96.1 $\pm$0.4       \\
 SearchCreativeWork       &   96.7 & 92.9 $\pm$1.0 & 94.2 $\pm$0.9 & 93.9 $\pm$1.0   & 94.2 $\pm$1.0       & 94.1 $\pm$1.1 & 94.6 $\pm$0.8  & 95.0 $\pm$0.5       \\
 SearchScreeningEvent     &   96.7 & 92.3 $\pm$0.8 & 94.0 $\pm$0.6 & 94.0 $\pm$0.6   & 93.5 $\pm$0.6       & 94.2 $\pm$0.4 & 94.8 $\pm$0.7  & 95.3 $\pm$0.3       \\
\hline
 Average                  &   96.7 & 93.1 $\pm$0.2 & 94.3 $\pm$0.2 & 94.2 $\pm$0.2   & 94.3 $\pm$0.3       & 94.5 $\pm$0.3 & 94.7 $\pm$0.3  & \textbf{95.0} $\pm$0.2       \\
\hline
\end{tabular}
}
\caption{SNIPS New-Intent few-shot results on \textbf{Global ST F1 Score}.}
\label{tab:snips-results-overall-ner}
\end{subtable}
\caption{Our results on SNIPS for the Global metrics, showing that the gains for Local metrics shown
in Tables \ref{tab:snips-results-new-ic} and \ref{tab:snips-results-new-ner} do not cause harm to the system overall.
See Section \ref{sub:snips_exp} for details.}
\end{table*}

\subsection{SNIPS Results on Global Metrics: NIFS Label Names Only (LNO)}
\label{sec:results_on_global_metrics_nifs_lno}

We show Global metrics for SNIPS in the NIFS-LNO setting
(New-Intent Few-Shot Label Names Only, Section \ref{sec:nifs_lno})
in Tables \ref{tab:snips-nifs-lno-global-ic} and \ref{tab:snips-nifs-lno-global-st}
for Intent Accuracy and Slot F1 Score, respectively.
These correspond to the Local metrics show in Tables \ref{tab:snips_lno_ic_new} for Local IC and \ref{tab:snips_lno_ner_new} for Local ST.
The numbers for ``\textsc{Linguist} (via s10 LNO)'' are only a few points behind ``s10'',
indicating that even when no real data is available for the novel intent,
\textsc{Linguist} generated data can provide some support for the new intent,
bringing the overall system performance close to where it would be with 10 real examples for that new intent.

\begin{table}[!htb]
\footnotesize
    \begin{subtable}{.5\linewidth}
      \centering
\begin{tabular}{l | c | c }
\hline
 Modified Intent / Data   &   s10      &   \thead{\textsc{Linguist}\\(via s10 LNO)}   \\
\hline
 AddToPlaylist            & 98.9 $\pm$ 0.3 & 94.5 $\pm$ 2.8              \\
 BookRestaurant           & 98.2 $\pm$ 0.3 & 98.0 $\pm$ 0.7              \\
 GetWeather               & 98.9 $\pm$ 0.1 & 99.1 $\pm$ 0.1              \\
 PlayMusic                & 96.1 $\pm$ 1.0   & 97.3 $\pm$ 0.5              \\
 RateBook                 & 98.9 $\pm$ 0.1 & 98.9 $\pm$ 0.1              \\
 SearchCreativeWork       & 96.5 $\pm$ 1.2 & 94.8 $\pm$ 0.8              \\
 SearchScreeningEvent     & 96.3 $\pm$ 1.1 & 92.6 $\pm$ 1.3              \\
\hline
 Average                  & 97.7 $\pm$ 0.2 & 96.5 $\pm$ 0.4              \\
\hline
\end{tabular}
\caption{SNIPS NIFS-LNO results on \textbf{Global Intent Accuracy}.}
    \label{tab:snips-nifs-lno-global-ic}
    \end{subtable}
    \begin{subtable}{.5\linewidth}
      \centering
\begin{tabular}{l | c | c }
\hline
 Modified Intent / Data   &   s10      &   \thead{\textsc{Linguist}\\(via s10 LNO)}   \\
\hline
 AddToPlaylist            & 94.2 $\pm$ 0.5 & 88.8 $\pm$ 1.9              \\
 BookRestaurant           & 94.1 $\pm$ 0.6 & 93.1 $\pm$ 0.9              \\
 GetWeather               & 94.9 $\pm$ 0.7 & 93.1 $\pm$ 0.6              \\
 PlayMusic                & 92.9 $\pm$ 0.3 & 92.6 $\pm$ 0.5              \\
 RateBook                 & 95.8 $\pm$ 0.4 & 88.1 $\pm$ 1.1              \\
 SearchCreativeWork       & 94.2 $\pm$ 0.9 & 93.1 $\pm$ 0.5              \\
 SearchScreeningEvent     & 94.0 $\pm$ 0.6 & 90.0 $\pm$ 0.8              \\
\hline
 Average                  & 94.3 $\pm$ 0.2 & 91.2 $\pm$ 0.4              \\
\hline
\end{tabular}
\caption{SNIPS NIFS-LNO results on \textbf{Global Slot F1 Score}.}
\label{tab:snips-nifs-lno-global-st}
\end{subtable} 
\caption{Global metrics (Intent Accuracy, (a), left; Slot F1 Score, (b), right) for SNIPS in the NIFS-LNO setting
(New-Intent Few-Shot Label Names Only, Section \ref{sec:nifs_lno}).
The numbers for ``s10'' are copied from
Tables \ref{tab:snips-results-overall-ic} and \ref{tab:snips-results-overall-ner},
for IC and ST, respecitvely.}
\label{tab:snips-nifs-lno-global}
\end{table}

\section{Intent Bleeding Case Study} % (fold)
\label{sec:intent_bleeding_case_study}

As described in Ex2 \citep{lee21ex2}, we also observed intent ``bleeding'',
where the model would produce outputs like one of the fine-tuning intents,
despite the prompt and examples being from a novel intent. We noticed this particularly
strongly when generating for \texttt{AddToPlaylist} intent on SNIPS,
where the model had a strong tendency to return utterances starting with ``play'',
which overlaps with the closely related \texttt{PlayMusicIntent} from fine-tuning.
Consequently, this harmed IC results (Table \ref{tab:snips-results-new-ic})
for this intent.
A very simple fix is to use ``n-gram blocking'', where the model is prevented from generating
phrases like ``play'', ``I want to play'', etc. during generation.
We found that this mitigates the issue, and we can get near 100\% on Intent Recall for AddToPlaylist.
However custom designing which n-grams to block requires effort from human experts,
and does not scale to a large number of new intents and languages,
so in future work, we would like to explore more automated and scalable solutions.

\section{Generation Hyperparameters} % (fold)
\label{sec:generation_hyperparameters}

For all three datasets, we use top-k sampling \citep{fan18nsg}.
For SNIPS, we use top\_k=50, temperature=0.3, and produce 100 outputs per input.
For mATIS++, we use top\_k=50, temperature=0.3, and produce 10 outputs per input.
For the internal dataset, we top\_k=20, temperature=1.0, and produce 20 outputs per input.
We observe that when \textsc{Linguist} is trained on the much larger internal dataset compared to SNIPS,
it produces less noisy outputs.
Thus, we allow a higher temperature of 1.0.
We use the same settings for all intents.

We also benchmarked beam search and nucleus sampling \citep{holtzman19degen} for generation,
and found both to perform worse overall on the internal
datasets and on SNIPS compared to top-k sampling.

% section generation_hyperparameters (end)

\section{Filtering ICLM Outputs} % (fold)
\label{sec:filtering_iclm_outputs}

We discard any outputs containing the \texttt{<unk>} token, which happens less than 1\% of the time.
The number of outputs (after de-duplication) are reported in Table \ref{tab:num-outputs-iclm}. 

\begin{table}[h!]
\centering
\footnotesize
\begin{tabular}{lc}
\hline
 Modified Intent       &   Num outputs \\
\hline
 AddToPlaylist        &           296 \\
 BookRestaurant       &           347 \\
 GetWeather           &           322 \\
 PlayMusic            &           255 \\
 RateBook             &           288 \\
 SearchCreativeWork   &           295 \\
 SearchScreeningEvent &           273 \\
\hline
 Average              &           297 \\
\hline
\end{tabular}
\caption{The number of filtered and de-duplicated outputs from ICLM per intent.
All numbers are averaged across the five random seeds.}
\label{tab:num-outputs-iclm}
\end{table}

\section{Filtering BT-Small Outputs} % (fold)
\label{sec:filtering_bt_small}

The small model has a fair amount of noise in its outputs,
so we heuristically filter them,
discarding any which contain repeated bigrams such as
\texttt{play the song halo the song} and/or
any trigram of the same word such as \texttt{of of of}.
Success rate and number of outputs (after de-duplication) are reported in Table \ref{tab:bt-small-success-rate}.

\begin{table}[h!]
\centering
\footnotesize
\begin{tabular}{lrrr}
\hline
 Modified Intent      &   SuccessRate &   NumOutputs &   AvgNumSlots \\
\hline
AddToPlaylist        &          70.2 &         64 &           2.7 \\
BookRestaurant       &          72.8 &         73 &           3.2 \\
GetWeather           &          60.4 &         60 &           2.3 \\
PlayMusic            &          53.6 &         52 &           2.2 \\
RateBook             &          70.8 &         71 &           3.8 \\
SearchCreativeWork   &          41.6 &         42 &           1.8 \\
SearchScreeningEvent &          69.6 &         70 &           2.2 \\
\hline
Average              &          62.7 &         62 &           2.6 \\
\hline
\end{tabular}
\caption{For each intent, the Success Rate of Back-Translation with the Small model,
and Number of Generated Outputs, both averaged across
the five random seeds.
For reference, we also show the Average Number of Slots in the training data per intent.}
\label{tab:bt-small-success-rate}
\end{table}

\section{Filtering BT-5B outputs} % (fold)
\label{sec:filtering_bt_5b_outputs}

The Back-Translated text with the 5B model is significantly cleaner than with the smaller model,
so we do not apply any filtering on the output text itself.
We do heuristically discard any outputs where we suspect the augmented
utterance is missing a slot. Specifically, SimAlign in ArgMax mode only
returns alignments across words that have mutual argmax between source and target.
For any source word that is an entity tag (i.e., not ``O''), if it is not aligned to
an output word, then we consider the output invalid. For example, an input like
\texttt{rate this book 5 out of 6} with a Back-Translated output
\texttt{give this book a rating of 5} would typically have no output word aligned to the source
word ``6'' (\texttt{best\_rating} slot label), so the output would be discarded.

Success rate and number of outputs (after de-duplication) for BT-5B are reported in Table \ref{tab:bt-5b-success-rate}.

\begin{table}[h!]
\centering
\footnotesize
\begin{tabular}{lrrr}
\hline
 Modified Intent       &   SuccessRate &   NumOutputs &   AvgNumSlots \\
\hline
 AddToPlaylist        &          66.2 &        411 &           2.7 \\
 BookRestaurant       &          82.8 &        423 &           3.2 \\
 GetWeather           &          72.0 &        311 &           2.3 \\
 PlayMusic            &          89.0 &        455 &           2.2 \\
 RateBook             &          79.2 &        478 &           3.8 \\
 SearchCreativeWork   &          85.5 &        451 &           1.8 \\
 SearchScreeningEvent &          72.0 &        330 &           2.2 \\
\hline
 Average              &          78.1 &        408 &           2.6 \\
\hline
\end{tabular}
\caption{For each intent, the Success Rate of Back-Translation with the 5B model, and
the number of outputs, both averaged across the five random seeds.
For reference, we also show the Average Number of Slots in the training data per intent.}
\label{tab:bt-5b-success-rate}
\end{table}

\section{Filtering \textsc{Linguist} Outputs} % (fold)
\label{sec:filtering_linguist_outputs}

% Filtering
We apply heuristic filtering by discarding outputs which
meet any of the following criteria:
(1) copy one of the examples from the prompt verbatim;
(2) fail to follow the prompt instructions, by
not copying the instructed slot value or
by producing repeated, missing, extra, or malformed slot-tag numbers;
(3) produce the literal wildcard instruction \texttt{"*"};
or (4) produce a content word containing a punctuation character in the set of
\{\texttt{\_<>[]()\{\};}\}.\footnote{These characters do not appear in the text of
any of the original training data, so are considered to be generation mistakes.}

In Table \ref{tab:gen-success-rate}, we report the Success Rate as the portion of generated utterances
which remain after filtering, and show the total number of generated utterances per intent.
We observe a trend that success rate is generally lower when the prompt contains more slots,
which is intuitive as the generation task is more challenging and has more chances to make a mistake.
The success rates vary significantly by intent from 75.1 for BookRestaurant to 96.9 for GetWeather,
with an average of 87.4 across the 7 intents.

\begin{table}[h!]
\centering
\footnotesize
\begin{tabular}{lccc}
\hline
 Modified Intent       &   Success Rate &   \#Outputs &   Average \#Slots \\
\hline
 AddToPlaylist        &          95.1 &         1230 &           2.7 \\
 BookRestaurant       &          75.1 &         2124 &           3.2 \\
 GetWeather           &          96.9 &         1197 &           2.3 \\
 PlayMusic            &          82.3 &         622 &           2.2 \\
 RateBook             &          78.0 &         1729 &           3.8 \\
 SearchCreativeWork   &          91.3 &         1154 &           1.8 \\
 SearchScreeningEvent &          93.0 &         1370 &           2.2 \\
 \hline
 Average              &          87.4 &         1346 &           2.6 \\
 \hline
\end{tabular}

\caption{For each intent, the Success Rate of Generation,
and Number of Generated Outputs, both averaged across
the five random seeds.
For reference, we also show the Average Number of Slots in the training data per intent.}

\label{tab:gen-success-rate}
\end{table}

\section{SNIPS Dataset Details}
\label{sec:snips_dataset_details}

We retrieve the SNIPS dataset from \href{https://github.com/sonos/nlu-benchmark/tree/master/2017-06-custom-intent-engines}{https://github.com/sonos/nlu-benchmark/tree/master/2017-06-custom-intent-engines}.
For each intent, we use ``full'' training set file, e.g.
\texttt{AddToPlaylist/train\_AddToPlaylist\_full.json}
to split into Train and Development sets (described in Section \ref{ssub:snips_dataset}).
The validation data comes from the ``validate''
file for each intent, e.g.
\texttt{AddToPlaylist/validate\_AddToPlaylist.json}.

In \texttt{PlayMusic/train\_PlayMusic\_full.json}, as of the time of publishing,
there is a bug in row 461 (0-based), where we replace \texttt{"Pop Punk Perfection <non\_utf8\_chars>"} with \texttt{"Pop Punk Perfection"} before processing the dataset.

We provide in Table \ref{tab:snips_s10_row_ids} the row IDs (0-based) and md5sum of the training data subsets we use for the ``s10'' New-Intent Few-Shot (NIFS) setting,
described in Section \ref{ssub:new_intent_few_shot_nifs}.

\begin{table}[h]
\setlength{\tabcolsep}{1.5pt}
\footnotesize
\centering
\begin{tabular}{c|l|rrrrrrrrrr|r}
\hline
   Seed & \multicolumn{1}{c|}{Intent}               &   \multicolumn{10}{c|}{row IDs} & \multicolumn{1}{c}{md5sum}                           \\
\hline
      \multirow{7}{*}{0} & AddToPlaylist        &  81 & 271 & 314 &  495 &  561 &  636 &  856 & 1285 & 1615 & 1702 & ade55e42e481f83c6617298d300758d8 \\
       & BookRestaurant       & 122 & 438 & 574 &  739 &  950 & 1252 & 1401 & 1420 & 1578 & 1728 & 60f6c7e4af848f3c7cfaedb02bb2f058 \\
       & GetWeather           & 163 & 348 & 454 &  529 &  870 &  932 &  966 & 1286 & 1368 & 1766 & f37c6a4040e861d517e046719eb8da10 \\
       & PlayMusic            & 348 & 454 & 529 &  808 &  827 &  870 &  966 & 1286 & 1368 & 1766 & a9899270bcfc3de3985d9b7149176916 \\
       & RateBook             & 125 & 129 & 181 &  690 &  739 & 1100 & 1243 & 1250 & 1600 & 1658 & bcfa3063511c156738669da43f657284 \\
       & SearchCreativeWork   &  75 &  76 & 126 &  256 &  272 &  412 &  611 &  712 & 1216 & 1301 & c8db82b42da972e2f3fbba4fb343ca97 \\
       & SearchScreeningEvent &  76 & 117 & 236 &  261 &  411 &  785 &  919 & 1523 & 1856 & 1866 & a32972f750a077709d22bd062b3a10f7 \\
\hline
      \multirow{7}{*}{1} & AddToPlaylist        &  15 &  26 &  80 &   91 &  459 &  637 &  723 &  735 &  844 & 1306 & 8463a2c95d6104cea85e096b1c9abb0f \\
       & BookRestaurant       & 172 & 246 & 829 &  999 & 1061 & 1100 & 1203 & 1602 & 1717 & 1901 & 50cf89c558c1fd4387861968fbb8ea72 \\
       & GetWeather           & 155 & 466 & 857 &  957 & 1514 & 1673 & 1687 & 1748 & 1810 & 1930 & 4252f8c48f17062a003e73438254bdf8 \\
       & PlayMusic            & 155 & 466 & 857 &  957 & 1514 & 1683 & 1687 & 1748 & 1810 & 1930 & 94aa305f40be0d971b5fbf019fed0525 \\
       & RateBook             & 210 & 349 & 506 &  527 &  596 &  745 & 1174 & 1241 & 1295 & 1426 & 0b87d112f3073a31f6bb6b406e0d4f0c \\
       & SearchCreativeWork   & 209 & 348 & 506 &  528 &  600 &  750 & 1175 & 1241 & 1299 & 1434 & 329937eef631008e1029f53ee7368513 \\
       & SearchScreeningEvent & 210 & 522 & 660 &  862 &  880 &  951 &  983 & 1314 & 1766 & 1925 & cdb4eb65ce6742a3106222d1e2b04add \\
\hline
      \multirow{7}{*}{2} & AddToPlaylist        & 177 & 244 & 912 & 1044 & 1047 & 1218 & 1306 & 1374 & 1423 & 1541 & f718676eafd6e2285ab11b2e35359aee \\
       & BookRestaurant       & 252 & 469 & 555 &  849 &  969 & 1053 & 1113 & 1324 & 1570 & 1800 & 269ad74c558d8be458d1ee9dff612e21 \\
       & GetWeather           &  16 &  40 & 345 &  384 &  611 &  694 &  735 & 1071 & 1128 & 1587 & 8a4d66e79fc26c900c20baf99c7f429d \\
       & PlayMusic            &  16 &  40 & 345 &  611 &  735 & 1071 & 1094 & 1128 & 1587 & 1840 & 45b22040849978b9590509ff53553661 \\
       & RateBook             & 561 & 840 & 937 & 1156 & 1234 & 1246 & 1314 & 1383 & 1401 & 1719 & 65b6a5bbc8ca80f4da4418362c16d299 \\
       & SearchCreativeWork   & 263 & 402 & 515 &  598 &  819 &  877 & 1116 & 1296 & 1657 & 1791 & 8d4f37ebd8f93e23a48f4ae60191303c \\
       & SearchScreeningEvent & 562 & 844 & 941 & 1160 & 1241 & 1253 & 1389 & 1393 & 1457 & 1635 & 4d4193a6a724b07a090bb483db8baded \\
\hline
      \multirow{7}{*}{3} & AddToPlaylist        & 381 & 491 & 885 &  909 & 1341 & 1451 & 1459 & 1580 & 1778 & 1873 & 160fadf3bf9b76ce0af603992ae3a025 \\
       & BookRestaurant       &  42 & 245 & 570 &  702 & 1059 & 1227 & 1283 & 1330 & 1465 & 1887 & 319bd8b3fad9666ae1ebc4a3fa4bb9de \\
       & GetWeather           &  90 & 127 & 502 &  522 &  759 &  910 &  957 & 1013 & 1242 & 1337 & 2188dd29d307e6200201eb0936b16a88 \\
       & PlayMusic            & 127 & 502 & 522 &  620 &  759 &  910 &  957 & 1093 & 1242 & 1840 & 3b1bb6d63a58ebe40b697aa1a7d9f1d7 \\
       & RateBook             &  42 &  70 & 372 &  447 &  768 & 1180 & 1594 & 1705 & 1838 & 1932 & 76fe7286f19552b820f0b9e4061bfe80 \\
       & SearchCreativeWork   &  42 &  70 & 372 &  447 &  772 & 1182 & 1207 & 1600 & 1711 & 1766 & 76de043a76b2c96d7e782ca890fb24d7 \\
       & SearchScreeningEvent &  42 &  70 & 179 &  274 &  454 &  889 &  957 & 1058 & 1061 & 1256 & 6651011e32bbadaf55e169251d1f3a31 \\
\hline
      \multirow{7}{*}{4} & AddToPlaylist        &  26 &  58 & 276 &  328 &  403 &  574 &  834 & 1069 & 1644 & 1891 & a2ce834b5ca753d67f88ff42530568f0 \\
       & BookRestaurant       & 228 & 270 & 519 &  946 & 1361 & 1482 & 1508 & 1832 & 1927 & 1936 & d3d8c652313485e537b56e3279dddf80 \\
       & GetWeather           &  11 & 213 & 371 &  442 &  948 & 1040 & 1140 & 1280 & 1659 & 1835 & b002ada961f3af1e95f34c46e2c3d047 \\
       & PlayMusic            &  11 & 213 & 277 &  371 &  442 &  948 & 1140 & 1280 & 1691 & 1835 & 81ed81653e9b7f8dbc0f05f38a97ecbb \\
       & RateBook             &  58 & 128 & 208 &  815 &  876 &  891 &  941 & 1772 & 1784 & 1879 & bce0c7d2d5b70a26378022ce3dba98fd \\
       & SearchCreativeWork   &  58 & 127 & 207 &  817 &  894 &  943 & 1640 & 1699 & 1788 & 1878 & 54caeb85f764adac5856dd006d819418 \\
       & SearchScreeningEvent &  58 & 128 & 596 &  894 &  952 &  962 & 1140 & 1365 & 1693 & 1928 & 767a9b5e0c83b4d043c74d57dcbba12e \\
\hline
\end{tabular}
\caption{The Row IDs (0-based) used for the ``s10'' splits of the SNIPS dataset.}
\label{tab:snips_s10_row_ids}
\end{table}

\section{SemER Metric} % (fold)
\label{sec:semer_metric}

For the internal IC+ST benchmark
(Sections \ref{ssub:internal_dataset}, 
\ref{ssub:metrics_for_internal_benchmark}, and 
\ref{ssub:linguist_internal}),
we report on Semantic Error Rate (SemER) \citep{Su2018ARS}
which jointly evaluates Intent Classification and Slot Filling.
SemER is defined as follows: comparing a reference of tokens and their accompanying labels,
count each of of these operations:
(1) Correct slots, where the slot name and slot value is correctly identified,
(2) Deletion errors, where the slot name is present in the reference but not in the hypothesis,
(3) Insertion errors, where extraneous slot names are included in the hypothesis,
(4) Substitution errors, where slot names from the hypothesis are included but with an incorrect slot value.
Intent classification errors are substitution errors.
Then, apply Equation \ref{eq:semer} to compute the SemER.

\begin{equation}
\textrm{SemER} = \frac{\textrm{\# Del + \# Ins + \# Sub}}{\textrm{\# Cor + \# Del + \# Sub}}
\label{eq:semer}
\end{equation}

\end{document}